\documentclass[10pt,twocolumn,letterpaper]{article}

\usepackage{cvpr}              
\usepackage[T1]{fontenc}
\usepackage{multirow}
\usepackage{booktabs}
\usepackage{array}
\usepackage{placeins}
\usepackage[accsupp]{axessibility}
\definecolor{cvprblue}{rgb}{0.21,0.49,0.74}
\usepackage[pagebackref,breaklinks,colorlinks,allcolors=cvprblue]{hyperref}

\def\paperID{******}
\def\confName{CVPR}
\def\confYear{2026}

\title{Thermal is Always Wild: Characterizing and Addressing Challenges in Thermal-Only Novel View Synthesis}

\author{
M. Kerem Aydin$^{1}$ \quad
Vishwanath Saragadam$^{2}$ \quad
Emma Alexander$^{1}$ \\
$^{1}$Northwestern University \quad
$^{2}$University of California, Riverside \\
\small \url{https://nubivlab.github.io/wild_thermal}
}

\begin{document}
\maketitle
\begin{abstract}

Thermal cameras provide reliable visibility in darkness and adverse conditions, but thermal imagery remains significantly harder to use for novel view synthesis (NVS) than visible-light images.
This difficulty stems primarily from two characteristics of affordable thermal sensors.
First, thermal images have extremely low dynamic range, which weakens appearance cues and limits the gradients available for optimization.
Second, thermal data exhibit rapid frame-to-frame photometric fluctuations together with slow radiometric drift, both of which destabilize correspondence estimation and create high-frequency floater artifacts during view synthesis, particularly when no RGB guidance (beyond camera pose) is available.
Guided by these observations, we introduce a lightweight preprocessing and splatting pipeline that expands usable dynamic range and stabilizes per-frame photometry. 
Our approach achieves state-of-the-art performance across thermal-only NVS benchmarks, without requiring any dataset-specific tuning.

\end{abstract}    
\section{Introduction}
\label{sec:intro}

Thermal cameras capture long-wavelength radiation emitted and reflected by surfaces, revealing structure that is often invisible to RGB sensors. 
Thermal cameras function reliably under conditions that challenge visible-light imaging; such as darkness, fog, or smoke, and expose information related to temperature and material properties.
These characteristics make thermal imaging a valuable sensing modality for applications including autonomous driving \cite{bhadoriya2022vehicle, chen2019pedestrian}, environmental monitoring \cite{vadivambal2011applications}, infrastructure inspection \cite{martinez2006automatic}, robotics \cite{correa2012human} and search-and-rescue \cite{yeom2024thermal}. 
Extending these capabilities to 3D perception through novel view synthesis (NVS) would enable reliable scene reconstruction in settings where visible-light cameras fail.

Novel view synthesis (NVS) enables reconstructing scene geometry and appearance from posed image collections and has become a reliable 3D perception tool for RGB imagery, supporting applications in robotics \cite{wang2024nerfs}, autonomous driving \cite{he2024neural}, and AR/VR \cite{li2025radiance}.
Its effectiveness stems from the rich texture, stable photometry, and consistent multiview observations typically available in visible-light data.
These assumptions, however, do not hold for thermal imagery, making NVS substantially more challenging despite its benefits in low-visibility settings.

\begin{figure*}[ht!]
  \centering
  \includegraphics[width=.9\textwidth]{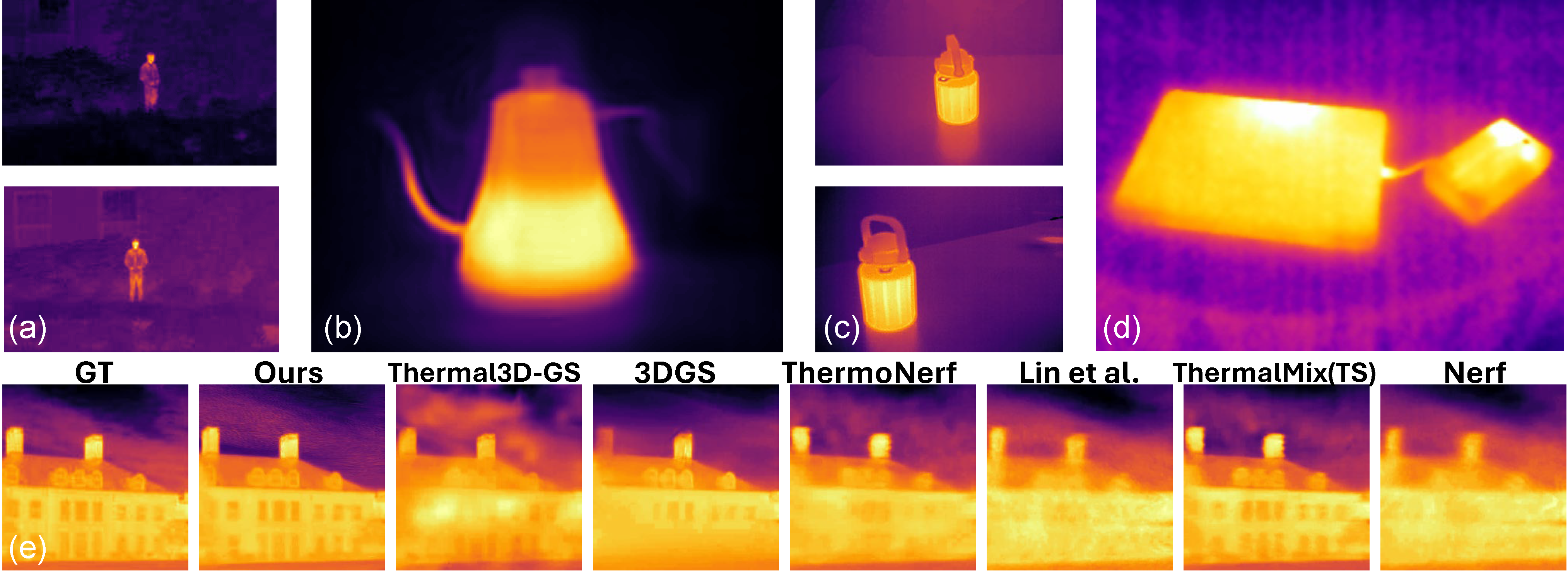}
    \caption{\textbf{Our method overcomes significant challenges in thermal images.} 
    Thermal data contain limited texture and lack multispectral cues, making correspondence estimation harder than in RGB. 
    They also exhibit sensor-specific degradations, including (a) frame-to-frame photometric inconsistency from sensor heating, (b) softened transitions between hot and cold regions characteristic of microbolometer sensors, (c) vignetting that produces viewpoint-dependent attenuation, and (d) fixed-pattern noise visible as structured artifacts.
    Our method explicitly stabilizes the photometry in (a), while the effects in (b–d) are mitigated in our SOTA reconstructions (e) through multiview consistency enabled by a novel embedding approach for learned appearance modeling. 
    }
   \label{fig:abb_ex}
\end{figure*}

Thermal images lack the chromatic and textural cues that support reliable correspondence in RGB and exhibit sensor-induced inconsistencies that disrupt geometric and photometric agreement across views.
These violations of multiview consistency make direct translation of RGB-based NVS pipelines unstable and often lead to geometric errors or radiometric artifacts.
Consequently, most prior work relies on paired RGB–thermal inputs \cite{lin2024thermalnerf, ozer2024exploring, hassan2024thermonerf, xu2024leveraging, zou2025tga, lu2024thermalgaussian}, using RGB to recover structure while treating thermal measurements as an auxiliary channel.
However, this dependence on visible-light imagery limits applicability in the very conditions where thermal sensing is most valuable, motivating the need for reliable thermal-only NVS.
\cref{fig:abb_ex} highlights the challenges in thermal data and illustrates our SOTA NVS performance.

This paper provides a rigorous framework for evaluating and improving thermal-only NVS by examining how thermal-specific degradations appear in real data.
To understand how these issues manifest in practice, we examine several public multiview thermal datasets and document their shared characteristics: low dynamic range, photometric fluctuations, slow radiometric drift, and limited texture across views.
Although the severity of these effects varies by dataset, they consistently reduce the stability of multiview correspondence and motivate the need for preprocessing steps that normalize thermal observations before reconstruction.
Guided by these observations, we develop a Gaussian-splatting pipeline tailored to the characteristics of thermal imagery.
Our approach integrates a lightweight preprocessing module that expands usable dynamic range and stabilizes per-frame photometry, followed by a splatting stage adapted to operate reliably under the reduced dynamic range, texture  and radiometric variability of thermal data.
Together, these components enable high-fidelity NVS from purely thermal inputs and improve performance across diverse datasets without dataset-specific tuning.

\section{Related Work}

\label{sec:dataset_analysis}
\begin{figure*}[ht!]
  \centering
  \includegraphics[width=.8\textwidth]{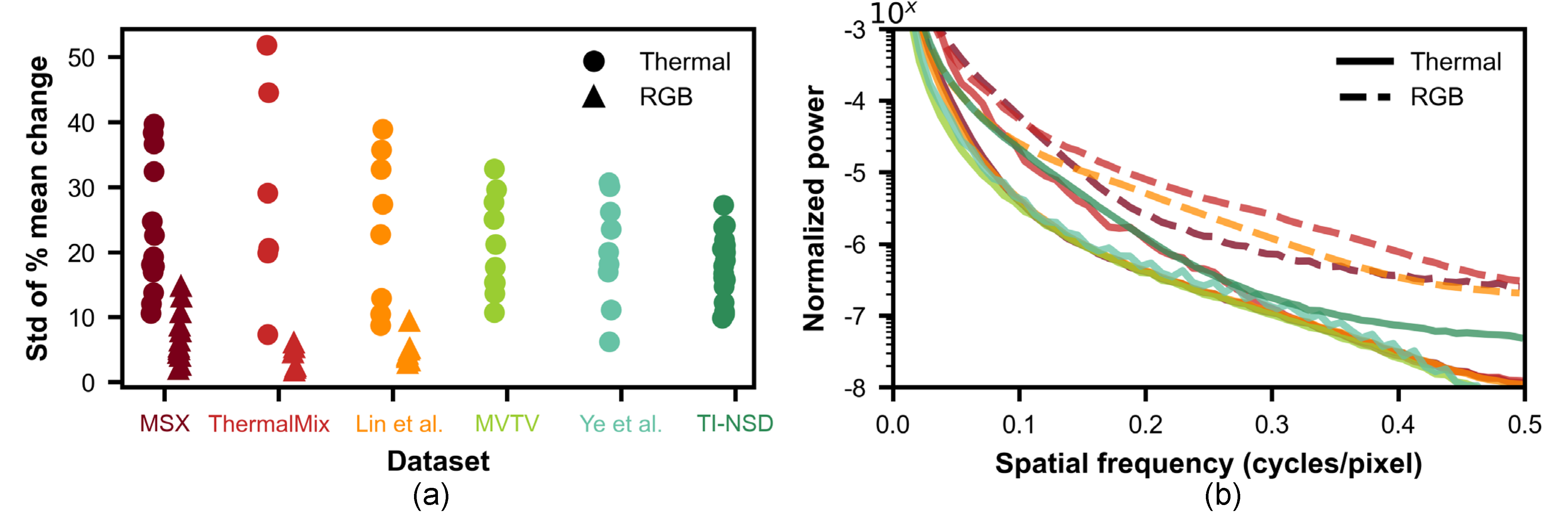}
    \caption{
        \textbf{Dataset-level radiometric and spatial-frequency characteristics.}
        (\textbf{a}) Standard deviation of the relative mean-intensity change $\Delta I_t$ across scenes. Thermal sequences show substantially larger radiometric fluctuations than RGB, with variation levels differing noticeably across datasets.  
        (\textbf{b}) Radially averaged power spectra, averaged per dataset. Thermal frames consistently exhibit reduced high-frequency energy compared to RGB; among thermal datasets, TI-NSD shows comparatively stronger high-frequency content.
}
       \label{fig:dataset_analysis}
\end{figure*}

\subsection{Novel View Synthesis}
Novel view synthesis (NVS) has advanced rapidly through representations that recover continuous scene structure from posed image collections.
Neural Radiance Fields (NeRF) made a major breakthrough by optimizing a volumetric radiance field to reproduce observed RGB views with high fidelity \cite{mildenhall2021nerf}.
Subsequent work improved robustness and efficiency through antialiasing \cite{barron2021mip} and fast multiresolution encodings \cite{muller2022instant}.
Explicit approaches further pushed efficiency by replacing neural volumetric fields with sparse voxel grids \cite{fridovich2022plenoxels} or with the anisotropic primitives used in 3D Gaussian Splatting \cite{kerbl20233dgaussiansplattingrealtime}.
Together, these advances illustrate the shift from slow neural radiance fields toward fast, explicit primitives that support high-quality NVS.

Standard NVS pipelines degrade when appearance varies across views, leading to drifting color estimates, incorrect density inference, and unstable geometry.
Several methods address these issues through appearance modeling, including per-frame embeddings for “in-the-wild” scenarios that adapt to illumination or style changes \cite{martin2021nerf}, or through joint pose refinement \cite{lin2021barf}.
Other approaches target extreme low-light conditions by training directly on noisy raw measurements \cite{mildenhall2022nerf}.
Recent extensions adapt Gaussian-splatting representations to uncontrolled photo collections by incorporating robust initialization, exposure normalization, or appearance conditioning \cite{kulhanek2024wildgaussians, xu2024wild, zhang2024gaussian, xu2024splatfacto, dahmani2024swag}.
Collectively, these works highlight the need for specialized techniques for NVS under appearance shifts, low-light noise, or pose uncertainty. These challenges are amplified in thermal imaging, where low contrast, radiometric drift, and imprecise poses further destabilize reconstruction. In this work, we extend the in-the-wild paradigm to thermal NVS, adapting appearance modeling to address these domain-specific challenges.

\subsection{Thermal Imaging}

Thermal cameras, and in particular the affordable microbolometer sensor-based images, suffer from strong artifacts that make NVS particularly challenging.
First, thermal images contain fixed pattern noise due to fabrication imperfections~\cite{saragadam2021thermal,he2018single}. Numerous hardware~\cite{olbrycht2015new} and algorithm-based~\cite{he2018single,liu2018shutterless,hardie2000scene,hardie2007map,wolf2016modeling, shin2022maximizing} solutions exist, but they often only partially address the fixed pattern noise.
Second, thermal cameras suffer from drifts due to internal heating, causing poor radiometric consistency~\cite{wolf2016modeling}. As we will see later, this has a debilitating effect on the quality of NVS images.
Finally, thermal images suffer from ghosting due to thermal inertia~\cite{ramanagopal2020pixel}, and lack texture, implying both pose estimation, and training the radiance fields is challenging.

\subsection{Thermal+RGB NVS}
RGB–thermal NVS methods leverage RGB imagery to compensate for the weak texture, radiometric drift, and pose instability characteristic of thermal sensors.
In \cite{ozer2024exploring}, the authors evaluate strategies for integrating thermal inputs into radiance-field pipelines, showing that explicitly adding a second thermal branch yields sharper reconstructions than fine-tuning or single-branch designs.
Lin et al. \cite{lin2024thermalnerf} takes a two-model approach, optimizing separate RGB and thermal NeRFs while enforcing cross-modal consistency through an $L_{1}$ density regularizer, allowing RGB to guide the geometry while preserving modality-specific appearance.
ThermoNeRF \cite{hassan2024thermonerf} applies a similar paired-modality formulation to architectural scenes, using aligned RGB–thermal images to learn density while predicting color and temperature through separate networks.
Additional work uses thermal measurements to improve reconstruction in low-light conditions \cite{xu2024leveraging, zou2025tga} or adapts Gaussian splatting to thermal inputs \cite{lu2024thermalgaussian}, and other approaches incorporate thermal cues to stabilize RGB NVS in smoke-filled environments\cite{jain2025smokeseer}.
Together, these methods demonstrate that RGB supervision significantly improves pose stability, geometry quality, and radiometric consistency for thermal data, but they require paired multimodal capture, motivating the need for thermal-only NVS methods.

\subsection{Thermal-Only NVS}
Thermal-only NVS aims to reconstruct geometry and appearance directly from thermal imagery, but prior work shows that weak texture, low dynamic range, and radiometric instability cause standard NeRF or 3DGS pipelines to struggle on thermal-only data ~\cite{lin2024thermalnerf, ozer2024exploring}. Here we consider the problem of view synthesis only, assuming that camera poses are known for all methods. In practice, estimating camera poses from thermal images in also difficult, requiring RGB images or other sensors (e.g., IMUs).
Thermal-NeRF learns radiance fields from a single infrared camera by introducing thermal mapping that normalizes intensity responses to 0-255 range and a structural patch constraint that stabilizes training in low-texture regions \cite{ye2024thermal}.
Thermal3D-GS adapts Gaussian splatting to thermal data through physics-inspired modeling and temperature-consistency constraints, and demonstrates improved reconstruction on a newly collected thermal multiview dataset \cite{chen2024thermal3d}.
Other thermal-only approaches explore emissive–residual Gaussian decompositions \cite{wen2025geometry}, degradation-aware radiance-field optimization for blurry or rolling-shutter thermal inputs \cite{carmichael2025trnerf}, thermal radiance prediction with physically motivated rendering \cite{ding2025exploring}, and segementation-based preprocessing schemes that enrich thermal signals for pose recovery \cite{zhong2024tex}.
Together, these works show that thermal-only NVS is feasible but remains strongly limited by low-dynamic-range, blur, and radiometric instability inherent to thermal sensors.%

\section{An Analysis of Multiview Thermal Datasets}

Multiview thermal datasets display a range of sensor behaviors that differ from those commonly encountered in RGB imagery. 
Because these characteristics influence how reliably an NVS pipeline can interpret thermal observations, it is important to examine how they appear in real sequences before introducing a reconstruction method.
In this section, we analyze several representative datasets to identify recurring properties of thermal multiview capture that are most relevant to downstream NVS performance.

We analyze six publicly available multiview thermal datasets used across recent thermal NVS studies: Lin et al.\cite{lin2024thermalnerf}, Ye et al.\cite{ye2024thermal}, MVTV\cite{xu2024leveraging}, MSX\cite{lu2024thermalgaussian}, ThermalMix\cite{ozer2024exploring}, and TI-NSD\cite{chen2024thermal3d}.
These datasets differ in sensor type, acquisition protocol, and scene content, spanning both uncooled microbolometers and cooled detectors, indoor and outdoor environments, and static and mobile capture setups.
This diversity allows us to separate dataset-specific artifacts from properties that consistently appear across thermal imagery.

\begin{figure*}
    \centering
    \includegraphics[width=.77\textwidth]{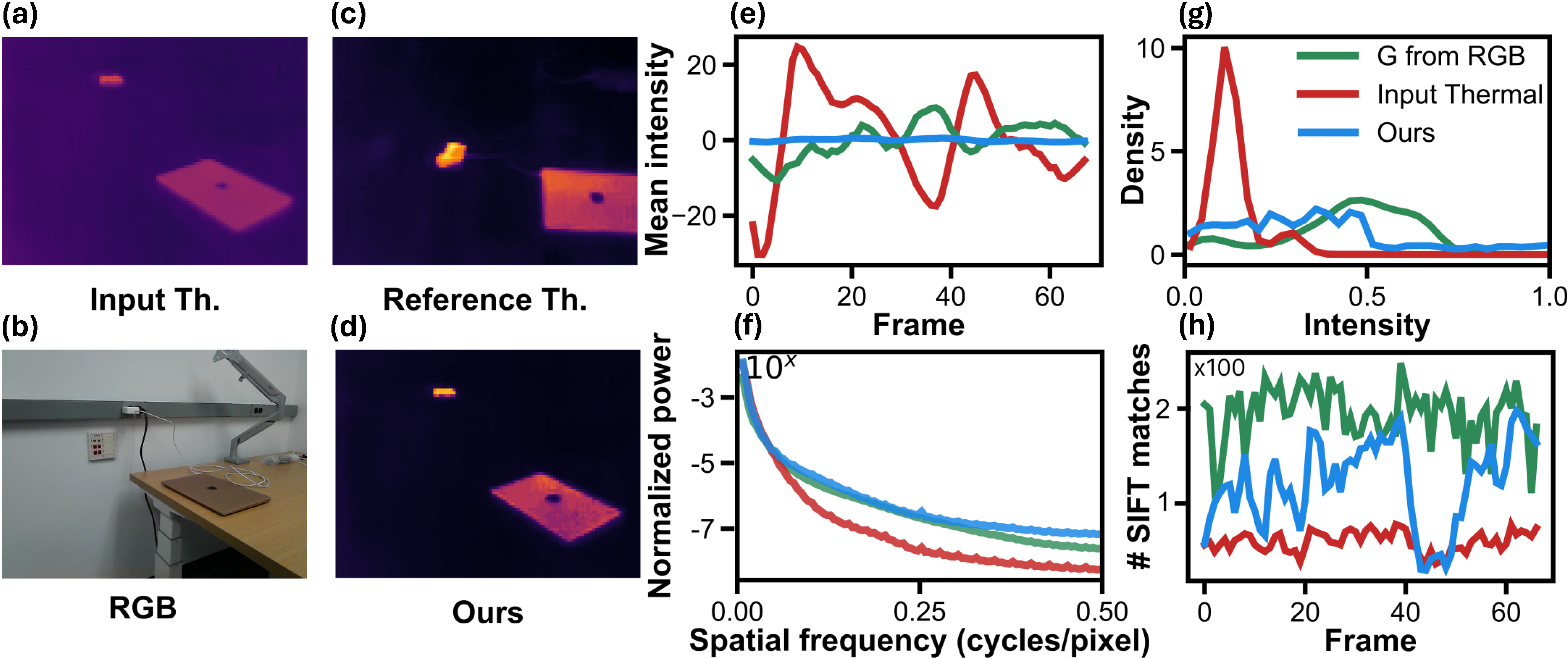}
    \caption{\textbf{Photometric stabilization and contrast enhancement} (a) An input thermal frame, with less contrast and texture than the corresponding RGB image (b). Notice the photometric drift when compared to reference frame (c), which shows the same scene at a different time point. 
    (d) Our invertible enhancement improves photometric inconsistency and image contrast. (e) Temporal mean intensity across frames showing reduced radiometric drift after stabilization. (f) Normalized spatial frequency spectrum with enhanced thermal data resembling RGB statistics. (g) Our preprocessing expands the effective dynamic range of thermal images, shown with pixel instensity distributions on a sample frame. (h) Improved structural consistency and stability, illustrated by an elevated number of tracked SIFT features per frame. See \cref{sec:sup:more_examples} for more examples.}
    \label{fig:methods_works}
\end{figure*}

We characterize dataset behavior using three diagnostics that capture photometric stability, spatial-frequency content, and effective dynamic range.
To assess radiometric stability, we measure the relative change in mean intensity across frames,
\begin{equation}
\Delta I_t = \frac{\mu_t - \bar{\mu}}{\bar{\mu}},
\end{equation}
where $\mu_t$ is the mean pixel value of frame~$t$ and $\bar{\mu}$ is the average intensity across the sequence.
Large fluctuations in $\Delta I_t$ indicate exposure drift or sensor-heating effects that disrupt brightness constancy and can destabilize correspondence estimation, often producing floater artifacts. Dataset-level trends appear in \cref{fig:dataset_analysis}a, with additional per-scene examples in the supplemental material (\cref{sec:sup:radiometric_consistency_examples}).
To analyze spatial-frequency characteristics, we compute radially averaged power spectra,
\begin{equation}
S_t(f) = \frac{1}{N_f} \sum_{(u,v): \|(u,v)\| \approx f} \big| \mathcal{F}(I_t)(u,v) \big|^2,
\end{equation}
which describe how image energy is distributed across frequencies. Higher frequencies correlate with texture, sharp edges, and sensor noise. 
Across datasets, thermal frames exhibit attenuated high-frequency responses due to the microbolometer’s smoothing behavior. This reduction in spatial detail weakens geometric and photometric cues that NVS methods typically rely on for stable multiview alignment.
Dataset-wide frequency trends are summarized in \cref{fig:dataset_analysis}b, with per-scene examples provided in the supplemental material (\cref{sec:sup:frequency_content_examples}).
Finally, we examine pixel-intensity histograms to evaluate effective dynamic range.
Thermal images often occupy a narrow portion of the available intensity space, leading to reduced contrast and weaker optimization gradients.
See \cref{fig:methods_works} for representative histogram behavior with additional examples at \cref{sec:sup:histogram_examples}.
Together, these diagnostics highlight the radiometric and contrast-related characteristics that shape thermal multiview data and identify the input properties that most challenge existing NVS pipelines.

In addition to these dominant trends, smaller artifacts such as vignetting and fixed-pattern noise further complicate multiview consistency. 
The following section introduces a lightweight preprocessing and splatting pipeline informed by these observations.

\section{Method}
\label{sec:method}

\begin{figure*}[ht!]
  \centering
  \includegraphics[width=
  .8\textwidth]{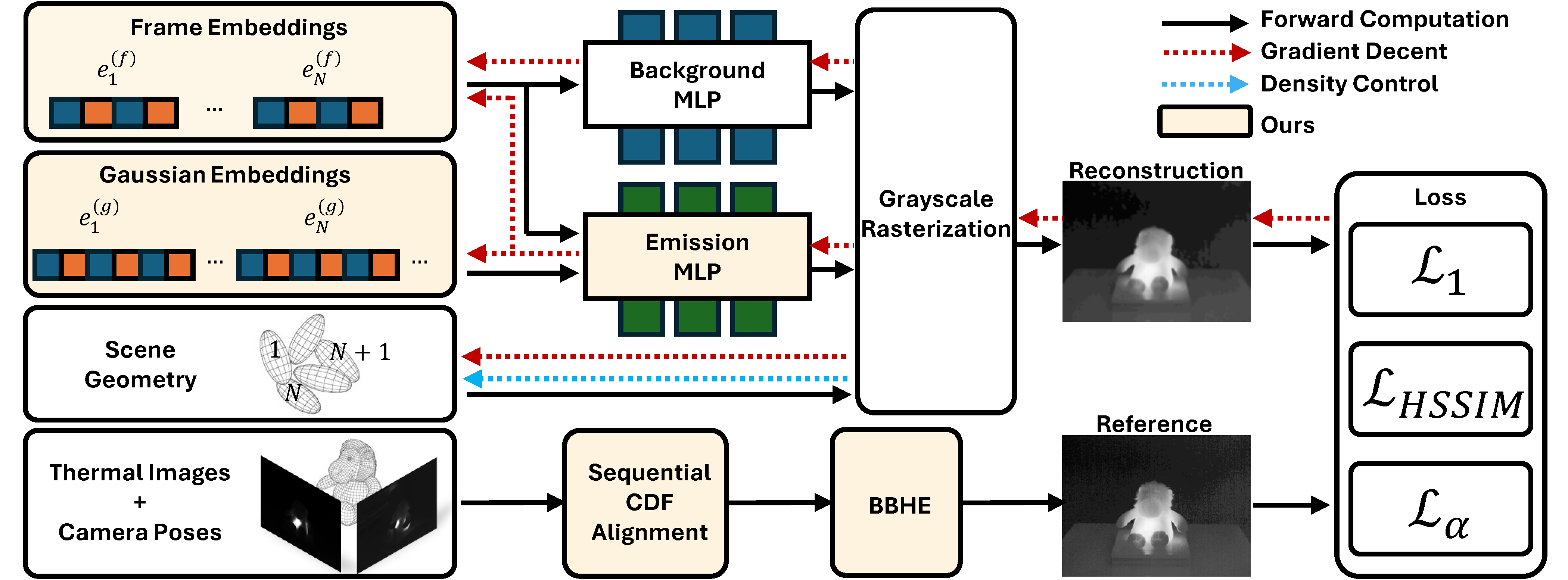}
  \caption{\textbf{Our pipeline.} Given thermal frames and camera poses, we first stabilize the inputs to ensure consistent training data across views (bottom). The frames are  modeled with a novel combination of per-Gaussian embeddings, which encode spatial appearance, per-frame embeddings, which capture residual temporal artifacts, and a physics-restricted parameter set (grayscale with no spherical harmonics) that stabilizes learning. These components jointly enable consistent thermal reconstruction while preserving fine geometric and intensity details.}
   \label{fig:pipeline}
\end{figure*}

Our approach consists of two components: a contrast–stabilizing preprocessing stage and an adaptation of 3D Gaussian Splatting (3DGS) to the thermal domain. 
The preprocessing normalizes temporal radiometry and enhances effective contrast, while the splatting stage models thermal appearance using a scalar emission representation conditioned on per-frame and per-Gaussian embeddings (Fig.~\ref{fig:pipeline}).

\subsection{Photometric Stabilization \& Enhancement}
\label{sec:method:preprocessing}
Thermal sequences exhibit frame-to-frame drift and low effective dynamic range (Sec.~\ref{sec:dataset_analysis}).
We address these effects using our two-step, monotonic transformation composed of sequential histogram alignment and brightness-preserving bi-histogram equalization (BBHE). 

Let $I_t$ denote the $t$-th frame, and let $x \in [0,1]$ denote a normalized pixel intensity.
We maintain an exponentially averaged reference Cumulative Distribution Function (CDF) $F_t^{*}$,
\begin{equation}
F_t^{*}(x)= (1-\alpha)F_{t-1}^{*}(x)+\alpha F_t(x),
\end{equation}
where $F_t$ is the CDF of $I_t$ and $\alpha$ controls temporal smoothing. 
Each frame is stabilized by mapping its distribution to the reference:
\begin{equation}
I_t'(x)=(1-\beta)x+\beta\,F_t^{*-1}(F_t(x)),
\end{equation}
with $\beta \in [0,1]$ balancing identity and alignment.
This step suppresses photometric drift while adapting smoothly to gradual scene changes.
We apply BBHE to $I_t'$, splitting the histogram at its mean intensity $T_t^{\mu}$ and equalizing the lower and upper subranges independently:
\begin{equation}
\hat{I}_t(x)=
\begin{cases}
T_t^{L}(x), & x\le T_t^{\mu}, \\
T_t^{U}(x), & x> T_t^{\mu}.
\end{cases}
\end{equation}
Because both operations are monotonic and one-to-one, the overall transformation is analytically invertible and preserves a recoverable mapping to the original radiometric scale through a LUT.
See ~\cref{fig:methods_works} and \cref{sec:sup:more_examples} for examples of stabilized and contrast-enhanced frames.

\subsection{Modeling ``Wildness'' for Thermal Learning}
\label{sec:method:gaussian_splats}

We adapt 3D Gaussian Splatting to thermal data by simplifying the emission model and incorporating appearance embeddings to handle the residual inconsistencies observed after preprocessing.
In thermal imaging, emission is effectively isotropic and single-channel; therefore, instead of predicting RGB colors with spherical harmonics, each 3D Gaussian stores a single scalar emission value. 
This reduces the complexity of color modeling and aligns the model to the physics of thermal imaging but it also makes the reconstruction more sensitive to frame-dependent fluctuations.

\begin{table*}
    \centering
    \caption{\textbf{Thermal-only NVS comparison across six multiview datasets.} We report mean PSNR and SSIM across all scenes for six publicly available datasets. Our analysis of dataset difficulty predicts systematic trends in performance across methods. By designing a pipeline that addresses thermal data challenges directly, we deliver SOTA performance on thermal-only NVS.} 
    \label{fig:results_table}
    \includegraphics[width=.95\textwidth]{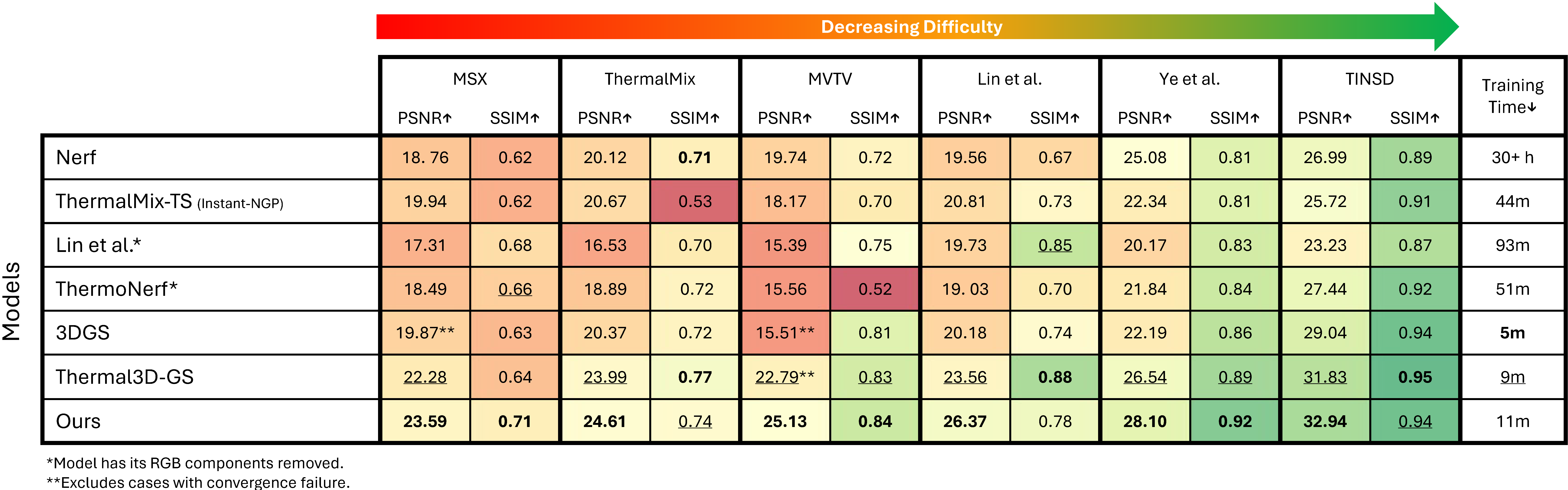}
\end{table*}

To address these residual inconsistencies, we adopt an embedding-based appearance formulation following the “3DGS-in-the-Wild’’ literature \cite{kulhanek2024wildgaussians, xu2024wild, zhang2024gaussian, xu2024splatfacto, dahmani2024swag}. 
Each Gaussian and each frame is assigned a small learnable embedding vector, $\mathbf{e}_i^{(g)}$ and $\mathbf{e}_t^{(f)}$, respectively, and a lightweight MLP maps these embeddings to a scalar emission,
\begin{equation}
c_i(t) = f_{\theta}\!\big(\mathbf{e}_i^{(g)},\,\mathbf{e}_t^{(f)}\big),
\end{equation}
which allows the model to absorb smooth frame-dependent variations caused by microbolometer heating, mild vignetting, or fixed-pattern noise without distorting the underlying geometry.
These emission values are then used in the usual 3DGS transmittance formulation,
\begin{equation}
\hat{I}_t(\mathbf{r}) = \sum_i T_i\,\alpha_i\,c_i(t),
\end{equation}
where $T_i$ denotes accumulated transmittance and $\alpha_i$ the opacity of each Gaussian along the ray.
During inference, fixing the frame embedding produces temporally stable reconstructions while leaving spatial detail unaffected.

As in recent RGB 3DGS algorithms, we use a background MLP to handle distant regions\cite{ye2024gaustudio, xu2024splatfacto}.
The rendered intensity blends foreground and background,
\begin{equation}
\tilde{I}_t(\mathbf{r}) = (1 - m(\mathbf{r}))\,\hat{I}_t(\mathbf{r}) + m(\mathbf{r})\,b_{\phi}(\mathbf{d},\,\mathbf{e}_t^{(f)}),
\end{equation}
where \(m(\mathbf{r})=\exp(-\sum_i \alpha_i)\) is the residual transmittance along the ray.

We train the model with a weighted sum of L1 error, heat-aware Structural Similarity Index Measure (SSIM)\cite{ye2024thermal}, and a background regularizer\cite{ye2024gaustudio}:
\begin{equation}
\mathcal{L} =
\lambda_{1}\,\mathcal{L}_{\text{L1}}
+ \lambda_{2}\,\mathcal{L}_{\text{HSSIM}}
+ \lambda_{3}\,\mathcal{L}_{\alpha}.
\end{equation}
The term $\mathcal{L}_{\text{HSSIM}}$ emphasizes thermal contrast and structure, while $\mathcal{L}_{\alpha}$ discourages floaters representing background regions.
\section{Results}
\label{sec:results}

\subsection{Implementation Details}
\label{sec:results:implementation}

Our system is implemented using the differentiable rasterization backend of GSplat~\cite{ye2025gsplat}, adapted for single-channel thermal imagery.
Each Gaussian stores a single emission coefficient without spherical harmonics, and the emission MLP uses three hidden layers of width 128 with ReLU activations followed by a linear output layer.
We train with the Adam optimizer and weight decay, using distinct learning rates for geometry, opacity, and appearance parameters.
All scenes are rendered at 1080p resolution; for quantitative evaluation, we crop and resize the outputs to match the original sensor resolution for pixel-wise alignment with ground truth.
Training runs for 30k iterations on an NVIDIA RTX A6000 GPU without learning-rate scheduling.
Gaussian centers are initialized from sparse COLMAP reconstructions, which recover stable but relatively sparse geometry in thermal data.

\subsection{Results and Comparison}
\label{sec:results:comparison}

\begin{figure*}
    \centering
    \includegraphics[width=.95\textwidth]{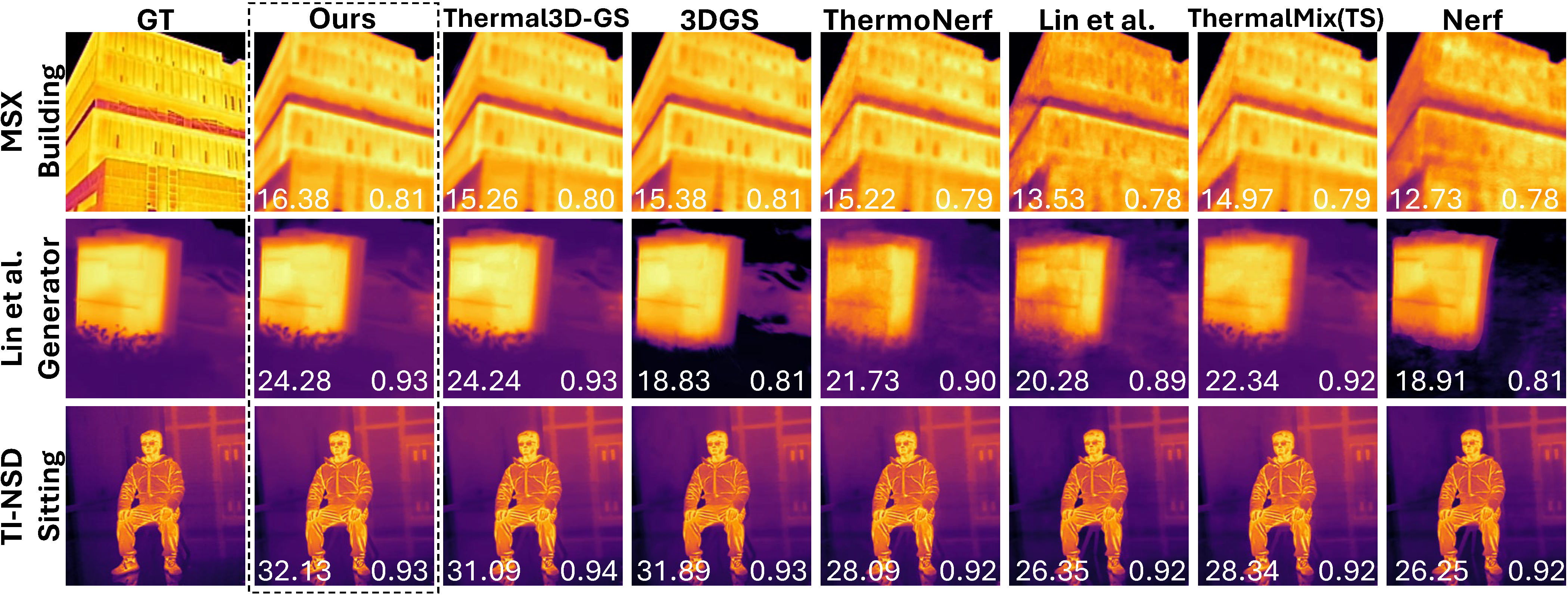}
    \caption{\textbf{Comparison to all methods.} Representative novel views, PSNR (bottom left, dB), and SSIM (bottom right) across datasets.
    Our method yields sharper boundaries and more stable background temperature on challenging scenes (top rows), while maintaining competitive quality on easier scenes (bottom rows).}
    \label{fig:results_images}
\end{figure*}

We compare our method against both general-purpose and thermal-specific neural reconstruction frameworks.
As standard references, we include NeRF~\cite{mildenhall2021nerf} and 3D Gaussian Splatting (3DGS)~\cite{kerbl20233dgaussiansplattingrealtime}.
To assess performance against recent thermal pipelines, we evaluate the ThermalMix-TS variant~\cite{ozer2024exploring}, derived from InstantNGP and designed for thermal sequences, and two cross-modal approaches, Lin et al.'s ThermalNeRF~\cite{lin2024thermalnerf} and ThermoNeRF~\cite{hassan2024thermonerf}.
Though the latter two originally rely on RGB–thermal supervision, we disable their RGB branches and cross-modal regularization losses to evaluate thermal-only performance.
Finally, we include Thermal3D-GS~\cite{chen2024thermal3d}, a thermal-only adaptation of Gaussian splatting that serves as the current state of the art.
All methods are trained under matching data splits, iteration counts, and hardware configuration to provide a consistent evaluation.
Quantitative results are summarized in \cref{fig:results_table}, and representative novel-view renders are shown in \cref{fig:abb_ex,fig:results_images}.

NeRF and ThermalMix-TS serve as baseline volumetric methods for thermal reconstruction.
Across all datasets in \cref{fig:results_table}, both models exhibit limited performance, with NeRF consistently ranking among the lowest in PSNR and SSIM due to its reliance on high-frequency RGB cues that are largely absent in thermal imagery.
As shown in \cref{fig:abb_ex,fig:results_images}, NeRF produces blurred surfaces and washed-out edges on the MSX Building and MVTV Mason scenes, and fails to recover background temperature on the Generator example.
ThermalMix-TS improves quantitative accuracy slightly, benefiting from faster optimization and better edge localization, yet it remains susceptible to noise and inconsistent global brightness.
While its hash-based representation enhances sharpness in small details, it also leads to unstable radiometric behavior across frames, leading to artifacts in video reconstructions. 
Overall, both methods reveal the inherent limitations of RGB-oriented volumetric frameworks when applied to thermal data.

\begin{figure}
    \centering
    \includegraphics[width=0.9\linewidth]{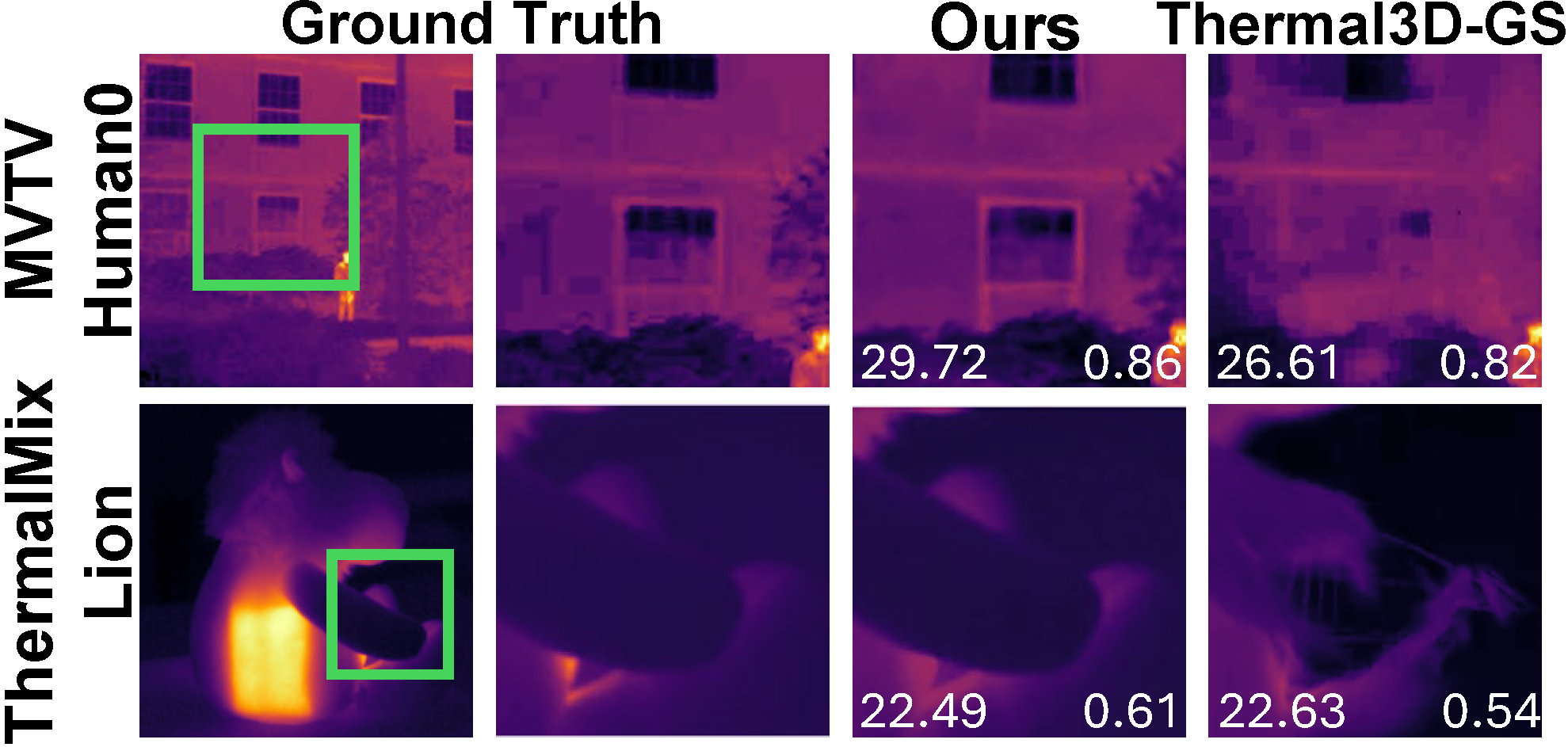}
    \caption{\textbf{Comparison to Thermal3D-GS.} Despite competitive metrics (PSNR in dB left, SSIM right), Thermal3D-GS exhibits significant failures in geometry and texture, while our method shows texture like the Human0 window, tree, and smooth wall, and faithfully reconstructs the Lion paw shape.}
    \label{fig:T3DGS}
\end{figure}

Both Lin et al. and ThermoNeRF were originally designed as cross-modal methods that jointly leverage RGB and thermal supervision.
In their original formulations, the RGB branch primarily drives geometric reconstruction, while the thermal channel acts as an auxiliary modality that paints radiometric information onto the recovered geometry and provides additional regularization.
When trained using thermal input alone, this coupling breaks down: both models lose geometric consistency and fail to recover fine-scale texture, producing oversmoothed surfaces and flattened temperature gradients.
As shown in \cref{fig:results_table}, their overall performance drops sharply across all datasets relative to thermal+RGB results previously reported in the literature.
Qualitatively (\cref{fig:results_images}), both methods achieve slightly better geometry than the baseline NeRF but still exhibit blurred edges and weak contrast.
In sequences with narrow-baseline training views, correspondence estimation becomes unstable, often leading to ghosting and duplicate surface artifacts in video reconstructions(see generator example in supplementary material).
These results demonstrate that cross-modal frameworks lose much of their effectiveness in a purely thermal setting, underscoring the need for reconstruction models designed from the ground up for thermal data.

\begin{figure*}
    \centering
    \includegraphics[width=0.92\textwidth]{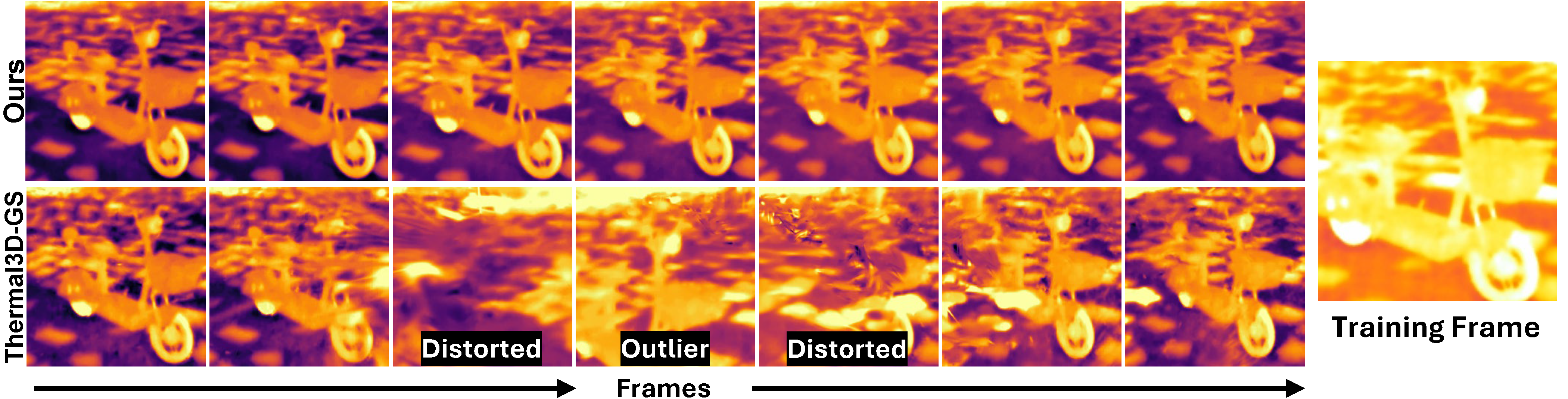}
    \caption{\textbf{Photometric consistency promotes high-quality reconstruction.} We render a smooth camera path for our method (top) and Thermal3D-GS (bottom).
    Thermal3D-GS produces bright floating structures that drift across frames and then assemble into a copy of the photometrically inconsistent training frame (right) when the viewpoint aligns.
    Our preprocessing and embedding-conditioned emission model handle this outlier without introducing floaters, yielding stable geometry throughout the trajectory.}
    \label{fig:floater_video_frames}
\end{figure*}

\begin{table*}[ht]
\centering
\setlength{\tabcolsep}{3.8pt}
\renewcommand{\arraystretch}{1.15}
\caption{\textbf{Ablation study.} We demonstrate that our preprocessing algorithm and “in-the-wild” architecture (3DGS + Emission MLP) independently improve 3DGS performance, and combining them yields superior results. Additional analysis comparing our method to traditional histogram equalization and evaluating each preprocessing step is provided in \cref{tab:sup_ablation}
}
\label{tab:ablation}
\newcommand{\fixedcol}[1]{\makebox[1.6cm][c]{#1}}
\footnotesize
\begin{tabular}{l|cc|cc|cc|cc|cc|cc|cc}
\toprule
\multirow{2}{*}{Method} &
\multicolumn{2}{c|}{MSX- Ebike} &
\multicolumn{2}{c|}{T. Mix - Lion} &
\multicolumn{2}{c|}{MVTV - Human} &
\multicolumn{2}{c|}{Lin et al. - Sink} &
\multicolumn{2}{c|}{Ye et al. - Seq.1} &
\multicolumn{2}{c|}{TINSD - Sitting} &
\multicolumn{2}{c}{Avg.} \\
 & PSNR & SSIM & PSNR & SSIM & PSNR & SSIM & PSNR & SSIM & PSNR & SSIM & PSNR & SSIM & PSNR & SSIM \\
\midrule
3DGS (Baseline) & 20.45 & 0.86 & 19.25 & 0.71 & 21.21 & 0.81 & 20.81 & 0.74 & 28.24 & 0.83 & \underline{29.51} & \textbf{0.88} & 22.25 & 0.81 \\
3DGS + Preprocessing & 22.79 & 0.86 & 24.11 & \textbf{0.82} & 22.01 & 0.84 & 23.71 & 0.78 & 29.90 & 0.85 & 22.42 & 0.85 & 23.01 & 0.83  \\
3DGS + Emission MLP & \underline{25.73} & \underline{0.89} & \underline{24.17} & \textbf{0.82} & \underline{24.22} & \underline{0.89} & \textbf{24.57} & \underline{0.83} & \underline{32.12} & \underline{0.89} & 25.98 & \underline{0.87} & \underline{24.93} & \underline{0.87} \\
\textbf{Ours} & \textbf{25.97} & \textbf{0.92} & \textbf{24.25} & \underline{0.81} & \textbf{26.18} & \textbf{0.90} & \underline{24.27} & \textbf{0.88} & \textbf{33.32} & \textbf{0.90} & \textbf{30.01} & \underline{0.87} & \textbf{26.14} & \textbf{0.88} \\
\bottomrule
\end{tabular}
\end{table*}

Among Gaussian-splatting methods, standard 3DGS provides a strong geometry prior but performs inconsistently on thermal data.
As reflected in \cref{fig:results_table}, it achieves reasonable SSIM but suffers from notable PSNR drops on ThermalMix and MVTV, and fails to converge on challenging MSX Iron Ingot/Landscape and MVTV Tree scenes.
These stability issues are  visible in \cref{fig:results_images} generator example, where the background collapses.
Thermal3D-GS improves stability through its ATF and TCM modules, yielding better numerical performance and generally sharper reconstructions than 3DGS. However, as shown in \cref{fig:T3DGS}, the method fails to capture accurate texture and geometry on scenes where we succeed, and does not converge on the MVTV Tree scene. 
These failures may be because their ATF module, originally designed to model atmospheric effects (and hence primarily attenuation of intensity), does not model the full variety of photometric inconsistencies that are inherent to thermal images.
When the sequence contains unusually bright frames, Thermal3D-GS underfits them, causing those observations to reappear as floating artifacts across a range of novel views.
\cref{fig:floater_video_frames} shows where a single photometrically inconsistent training frame induces floaters that gradually assemble into an entire training viewpoint in Thermal3D-GS, while our method remains stable.
Our method avoids these failures not by modeling atmospheric scattering, but by directly addressing the underlying issue: thermal sequences contain radiometric fluctuations that require a representation with higher flexibility than ATF’s attenuation-based formulation. 
By using photometrically stabilized inputs and an embedding-conditioned emission model with greater expressiveness, our approach can represent both bright and dark observations within a consistent radiometric space, preventing floaters and preserving geometry across views. 
This advantage is consistent across datasets, where we obtain the best or second-best PSNR/SSIM in \cref{fig:results_table} while maintaining stable behavior even on the most challenging sequences.
These differences are even clearer in the supplemental videos, where long camera paths reveal temporal stability and the absence of floaters more clearly than still images.
Although our per-scene training time ($\sim$11~min) is slightly longer than 3DGS (5~min) and Thermal3D-GS (9~min), the improvement in reconstruction fidelity and temporal stability represents a favorable quality--efficiency trade-off.

\subsection{Ablation Study}
\label{sec:results:ablation}

As shown in \cref{tab:ablation}, our preprocessing and emission modeling provide complementary improvements for thermal reconstruction.
Preprocessing alone provides modest but consistent gains by stabilizing frame-to-frame radiometric variation, improving average PSNR from 22.25 to 23.01~dB.
Our in-the-wild baseline, implemented with an emission MLP and embeddings, models residual temporal artifacts that fixed Gaussian colors cannot capture, raising average PSNR to 24.93~dB.
Combining both components produces the best overall performance (26.14~dB / 0.88~SSIM), as preprocessing also introduces stronger gradients that benefit the emission model.
Improvements are particularly strong on Human, Ebike, Lion, and Seq.1, while Sitting shows smaller gains due to the already high quality of the baseline reconstruction.
A detailed breakdown of preprocessing components and comparison to traditional histogram equalization are provided in \cref{sec:sup:ablation_extended}.

%
%
%
%

\section{Conclusion}
\label{sec:conclusion}
We presented a thermal NVS pipeline designed to address two persistent challenges in thermal imagery: strong frame-to-frame photometric inconsistencies and limited dynamic range.
Our approach combines a lightweight, invertible photometric stabilization and contrast enhancement stage with a thermal variant of 3D Gaussian Splatting.
The preprocessing aligns each frame to a temporally smooth reference distribution, producing stable, contrast-enhanced inputs for reconstruction.
On top of this, we adapt appearance-modeling components from in-the-wild NVS and incorporate a small background-emission MLP tailored to single-channel thermal data.
These components allow the model to absorb residual radiometric transients and aberrations.
Together, they make thermal NVS feasible without RGB supervision and improve reconstruction stability and fidelity under real-world conditions.

Despite these advantages, several limitations remain.
First, the photometric stabilization stage operates offline and is not jointly optimized with reconstruction, suggesting future end-to-end formulations that jointly learn radiometric normalization and geometry.
Pose estimation remains a bottleneck, as tools like COLMAP still underperform on stabilized thermal frames compared to RGB.
Developing more robust, thermal-aware pose estimation methods is an important direction for future work.

\FloatBarrier

\FloatBarrier
\pagebreak
\section*{Acknowledgements}

This work was supported in part by the National Science Foundation under Grant No. IIS-2106786-001, and by the University of California, Riverside 
Regents' Faculty Fellowship. We thank Yi-Chun Hung for feedback on the manuscript and Mark Sheinin for insightful early conversations.
{
    \small
    \bibliographystyle{ieeenat_fullname}
    \bibliography{main,refs_VS,refs_VS2}
}

\newcounter{si}
\setcounter{si}{1} 
\renewcommand\thesection{S\arabic{si}}
\newcounter{fi}
\setcounter{fi}{1} 
\renewcommand{\thefigure}{S\arabic{fi}}

\maketitlesupplementary
\setcounter{page}{1}

\newenvironment{supptable}[1][]{%
    \refstepcounter{suptable} 
    \renewcommand{\thetable}{S\arabic{suptable}} 
    \begin{table}[#1] 
}{
    \end{table}
}
\newenvironment{supptable*}[1][]{%
    \refstepcounter{suptable} 
    \renewcommand{\thetable}{S\arabic{suptable}} 
    \begin{table*}[#1] 
}{
    \end{table*}
}
\newcounter{suptable}
\setcounter{suptable}{0}

\section{More Examples on Thermal Image Characteristics}
\label{sec:sup:more_examples}

\subsection{Radiometric Inconsistency Examples}
\label{sec:sup:radiometric_consistency_examples}
\begin{figure}[ht!]
  \centering
  \includegraphics[width=0.8\linewidth]{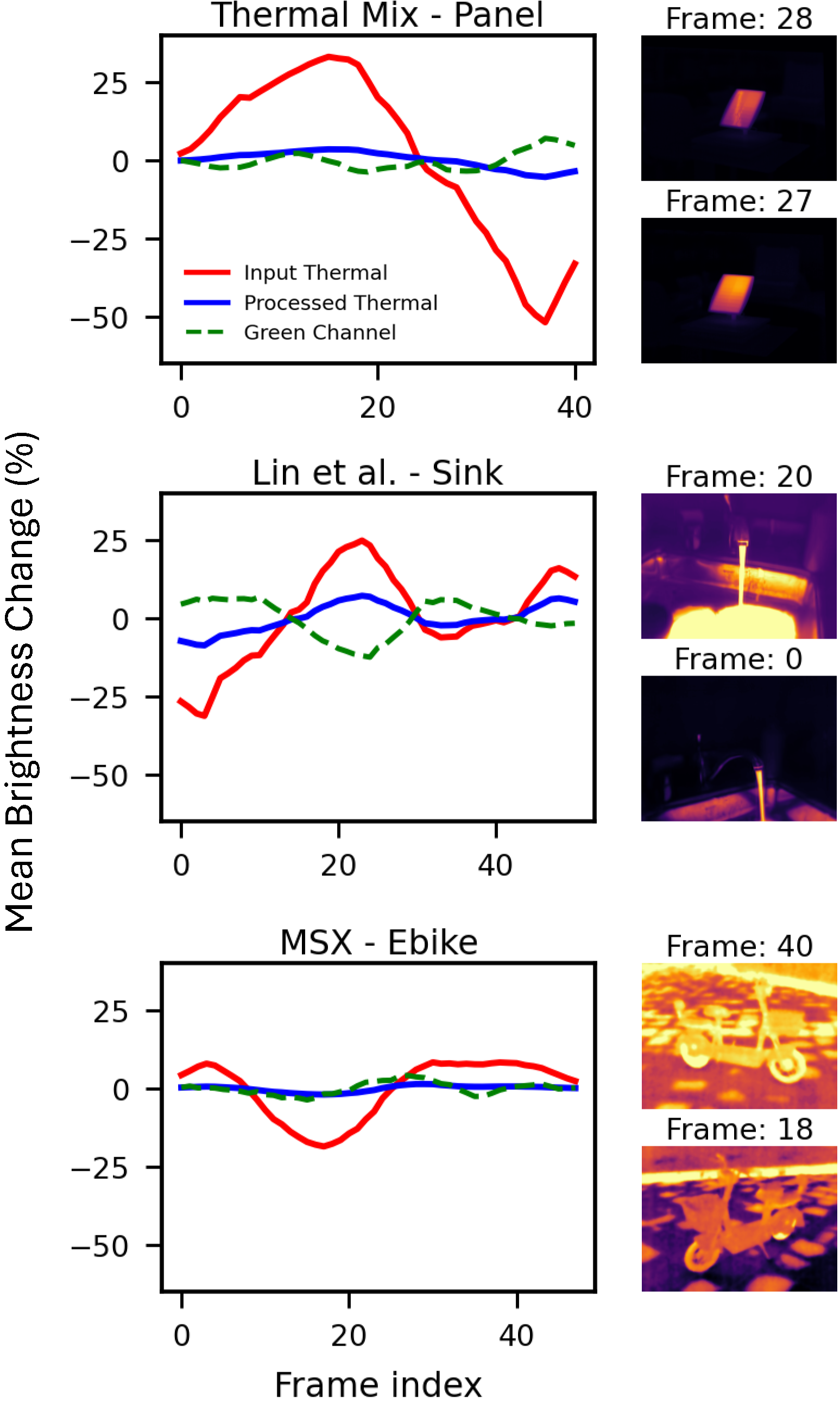}
   \caption{\textbf{Radiometric inconsistency across thermal datasets.} Thermal sequences often exhibit frame-to-frame fluctuations, generally influenced by ISP operations, as well as slower radiometric drift that can arise as the sensor gradually warms during capture. Both effects may appear even in static scenes and produce unstable brightness trajectories, as illustrated across the three datasets. These inconsistencies pose a significant challenge for NeRF and 3D Gaussian Splatting methods, which assume photometric consistency across views; variations in thermal intensity can weaken multi-view matching and lead to distorted geometry. In examples from Lin et al. and MSX, we observe global intensity shifts affecting nearly all pixel values, while the ThermalMix example shows a localized change in the foreground object. Our algorithm provides a simple yet effective way to suppress these trends and stabilize temporal behavior.}
   \label{fig:sup:pht_cons_ex}
   \stepcounter{fi}
\end{figure}
\FloatBarrier
\vfill
\pagebreak

\subsection{Frequency Content Examples}
\label{sec:sup:frequency_content_examples}
\begin{figure}[ht!]
  \centering
  \includegraphics[width=0.85\linewidth]{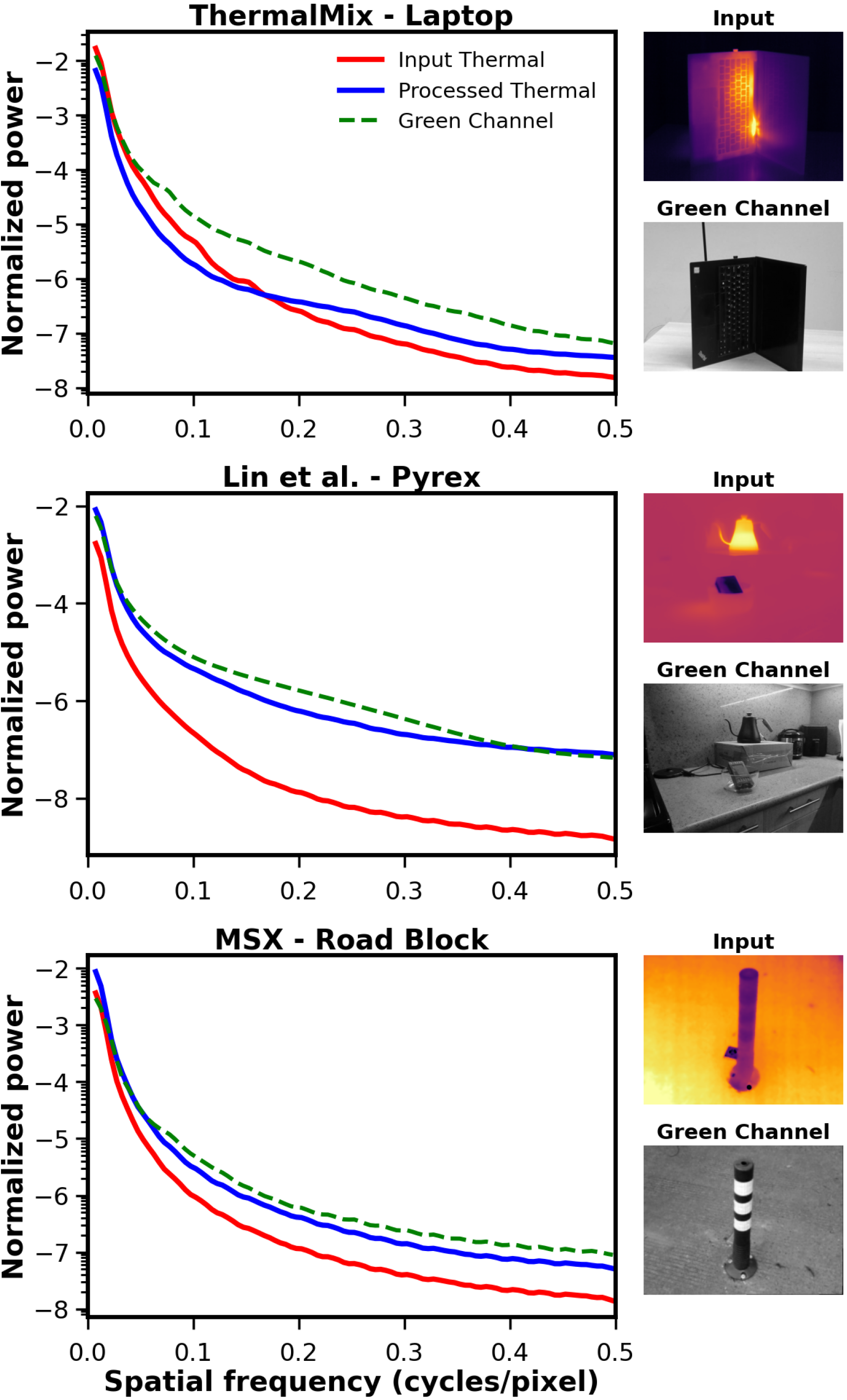}
    \caption{\textbf{Spatial frequency characteristics of thermal versus RGB images.}
    Thermal images generally exhibit weaker mid- and high-frequency components than RGB images. This is partly due to sensor-resolution limits, but also because heat diffuses across surfaces and through the surrounding air, producing the naturally smoother appearance typical of thermal scenes. As a result, fine texture and sharp edges are often diminished, reducing the high-frequency cues that NeRF and 3D Gaussian Splatting rely on for accurate geometry and appearance estimation. Our pipeline increases contrast and strengthens spatial gradients, making features more distinguishable while unavoidably amplifying noise and discretization artifacts. Even so, the enhanced gradients lead to more reliable COLMAP initializations (see \cref{fig:methods_works}) and provide stronger supervision for 3DGS, improving overall reconstruction quality.}
   \label{fig:sup:freq_ex}
   \stepcounter{fi}
\end{figure}
\FloatBarrier
\vfill
\pagebreak

\subsection{Histogram Examples}
\label{sec:sup:histogram_examples}
\begin{figure}[ht!]
  \centering
  \includegraphics[width=0.8\linewidth]{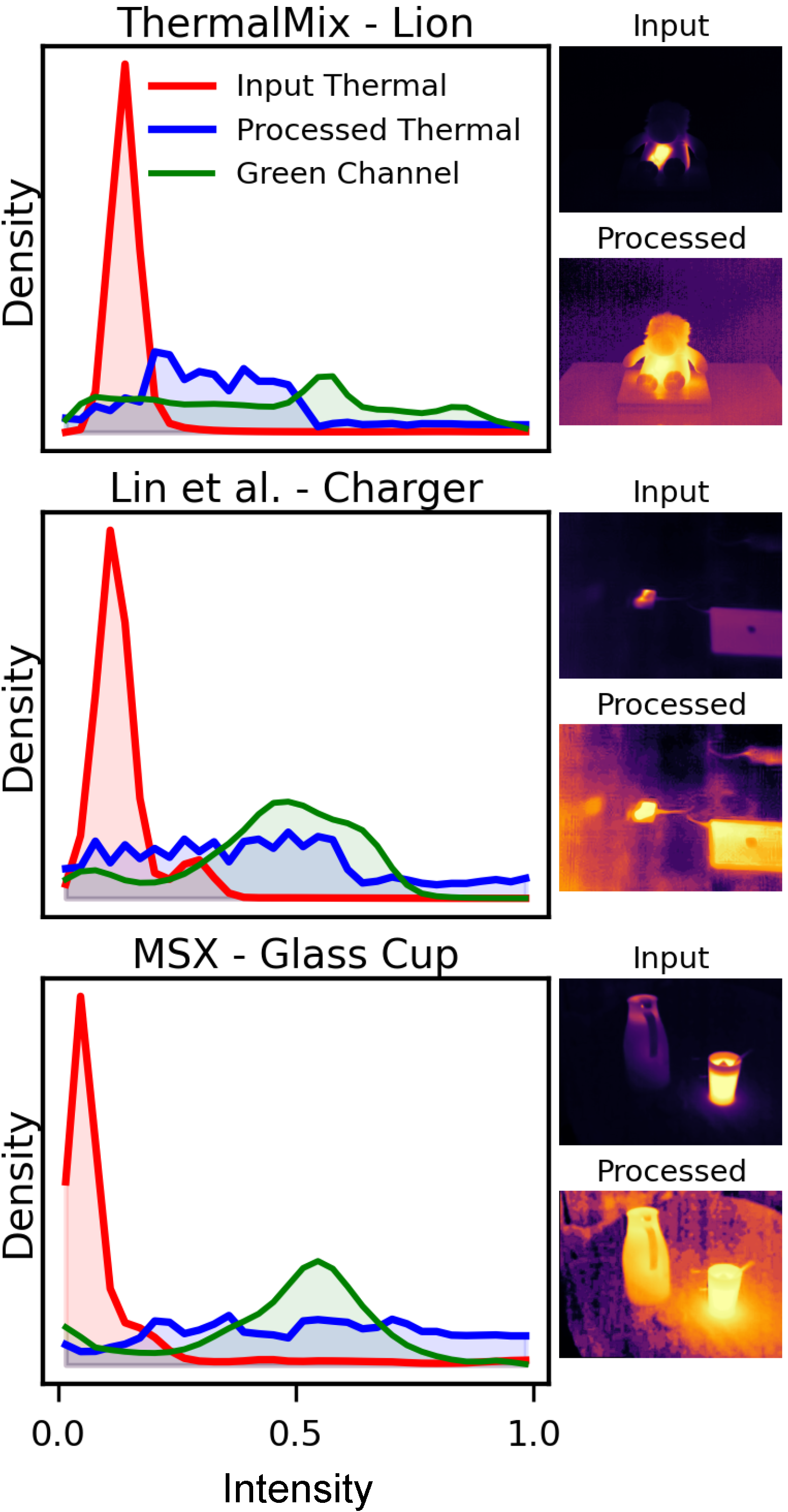}
    \caption{\textbf{Intensity distributions of thermal and RGB images.} Raw thermal images typically exhibit highly skewed histograms, where the majority of values accumulate near the lowest intensities. This narrow distribution limits contrast and reduces the visibility of meaningful variations in surface temperature, making it harder for NeRF and 3D Gaussian Splatting to extract useful spatial cues. Our pipeline expands the dynamic range and yields more spread-out intensity distributions, allowing important regions to stand out more clearly. Although this process can make noise more noticeable, the resulting images provide stronger supervision signals and improve reconstruction stability in 3DGS. Example input and processed frames are shown on the right.}
   \label{fig:sup:hist_ex}
   \stepcounter{fi}
\end{figure}
\FloatBarrier
\vfill
\pagebreak

\stepcounter{si}
\section{Local Effects of Preprocessing}

\begin{figure}
    \centering
    \includegraphics[width=0.9\columnwidth]{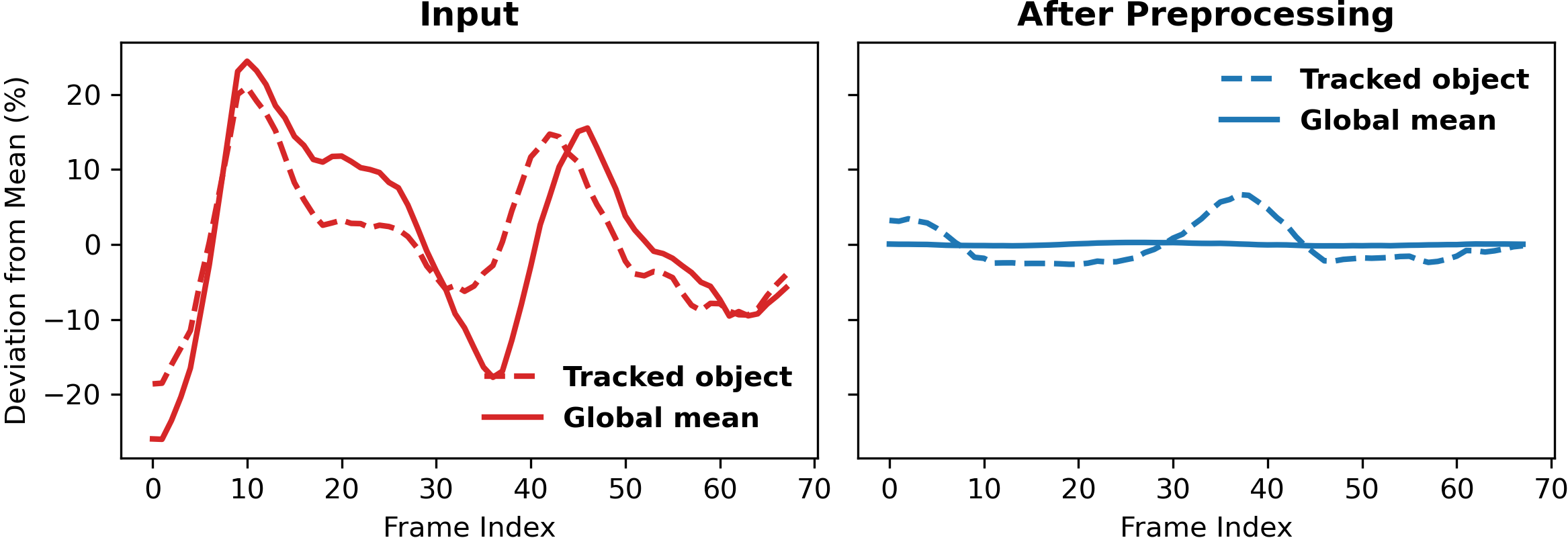}
    \caption{Revisit of Figure~3. Tracked object brightness and global mean before (left) and after (right) preprocessing; local fluctuations follow the global mean and are jointly attenuated.}
    \label{fig:sup:tracked_feature}
\end{figure}

To verify that our preprocessing stabilizes thermal intensity locally, we revisit example in \cref{fig:methods_works}. In \cref{fig:sup:tracked_feature}, we manually track the Apple logo on the laptop accross the frames and show how its intensity changes. While raw frames exhibit noticeable frame-to-frame fluctuations, our preprocessing significantly reduces this variability, producing a stable intensity trajectory for the tracked region.

\stepcounter{si}
\section{Ablation Study Extended}
\label{sec:sup:ablation_extended}

As shown in \cref{tab:sup_ablation}, each component of our system addresses a distinct failure mode of thermal reconstruction.
Photometric stabilization alone yields small but consistent gains by suppressing frame-to-frame brightness drift, whereas applying contrast enhancement in isolation is unreliable: it helps low-contrast scenes such as Lion and Sink but degrades scenes like Human0 and Ebike where gradients are already strong and contrast boosts amplify noise.
When stabilization and enhancement are combined, the effects become complementary, producing a more balanced improvement across datasets. 
The emission MLP provides the most substantial jump in performance by modeling residual radiometric transients that fixed Gaussian colors cannot capture, raising average PSNR from 23.01 to 24.93~dB.
Our full system achieves the best overall results (26.14~dB / 0.88~SSIM), with especially strong improvements on Human0, Ebike, Lion, and Seq.1 while the Sitting scene shows smaller gains due to already high image quality in the baseline reconstructions.
Additionally, in \cref{tab:sup_ablation}, we compare our preprocessing algorithm to traditional histogram equaalization, and independenly evaluate the effect of each preprocessing step.

\begin{supptable*}[ht]
\centering
\setlength{\tabcolsep}{3.8pt}
\renewcommand{\arraystretch}{1.15}
\caption{\textbf{Ablation study.} We demonstrate that our preprocessing steps (constrast enhancement as an improvement over traditional histogram equalization, in combination with photometric stabilization) improve 3DGS performance, but not to the level of our method. We then demonstrate the effect of per-frame embedding ablation, with and without pre-processing.}
\label{tab:sup_ablation}
\newcommand{\fixedcol}[1]{\makebox[1.6cm][c]{#1}}
\footnotesize
\begin{tabular}{l|cc|cc|cc|cc|cc|cc|cc}
\toprule
\multirow{2}{*}{Method} &
\multicolumn{2}{c|}{MSX- Ebike} &
\multicolumn{2}{c|}{T. Mix - Lion} &
\multicolumn{2}{c|}{MVTV - Human} &
\multicolumn{2}{c|}{Lin et al. - Sink} &
\multicolumn{2}{c|}{Ye et al. - Seq.1} &
\multicolumn{2}{c|}{TINSD - Sitting} &
\multicolumn{2}{c}{Avg.} \\
 & PSNR & SSIM & PSNR & SSIM & PSNR & SSIM & PSNR & SSIM & PSNR & SSIM & PSNR & SSIM & PSNR & SSIM \\
\midrule
3DGS (Baseline) & 20.45 & 0.86 & 19.25 & 0.71 & 21.21 & 0.81 & 20.81 & 0.74 & 28.24 & 0.83 & \underline{29.51} & \textbf{0.88} & 22.25 & 0.81 \\
3DGS + Hist. Eq. & 12.54 & 0.48 & 20.29 & 0.79 & 16.85 & 0.68 & 21.02 & 0.74 & 28.62 & 0.81 & 21.12 & 0.82 & 18.86 & 0.72\\
3DGS + Contrast Enh. & 14.79 & 0.57 & 23.97 & \underline{0.81} & 17.33 & 0.68 & 21.89 & 0.71 & 29.90 & 0.82 & 20.59 & 0.80 & 19.71 & 0.73 \\
3DGS + Photo.\ Stab. & 23.72 & 0.87 & 19.24 & 0.71 & 21.65 & 0.81 & 21.82 & 0.76 & 28.44 & 0.82 & 24.50 & 0.86 & 22.19 & 0.81 \\
3DGS + Stab + Enh  & 22.79 & 0.86 & 24.11 & \textbf{0.82} & 22.01 & 0.84 & 23.71 & 0.78 & 29.90 & 0.85 & 22.42 & 0.85 & 23.01 & 0.83  \\
3DGS + Emission MLP & \underline{25.73} & \underline{0.89} & \underline{24.17} & \textbf{0.82} & \underline{24.22} & \underline{0.89} & \textbf{24.57} & \underline{0.83} & \underline{32.12} & \underline{0.89} & 25.98 & \underline{0.87} & \underline{24.93} & \underline{0.87} \\
\textbf{Ours} & \textbf{25.97} & \textbf{0.92} & \textbf{24.25} & \underline{0.81} & \textbf{26.18} & \textbf{0.90} & \underline{24.27} & \textbf{0.88} & \textbf{33.32} & \textbf{0.90} & \textbf{30.01} & \underline{0.87} & \textbf{26.14} & \textbf{0.88} \\
\bottomrule
\end{tabular}
\end{supptable*}

\stepcounter{si}
\section{Hyperparameters and Weights}
\label{sec:sup:parameters}
Pre-trained weights are provided in our GitHub repository, which also contains the full training code: 

\small{\url{https://github.com/NUBIVlab/wild_thermal}}

\stepcounter{si}
\section{Video Results}
We include full video results in the supplementary materials and show zoomed-in insets for closer inspection. Since Thermal3D-GS is the current state of the art in thermal NVS, these insets focus on comparisons with that method. Note that because these synthesized views are not, generally, part of the test set, there is no ground truth, so large frames are our reconstructions.

To generate novel video frames, we move a virtual camera along pre-determined paths that traverse a subset of the training views. For methods that optimize camera poses, this sometimes produces minor deviations from the intended path, especially in scenes with very limited texture. 

\subsection{MVTV - Human0}
\begin{figure}[ht!]
  \centering
  \includegraphics[width=\linewidth]{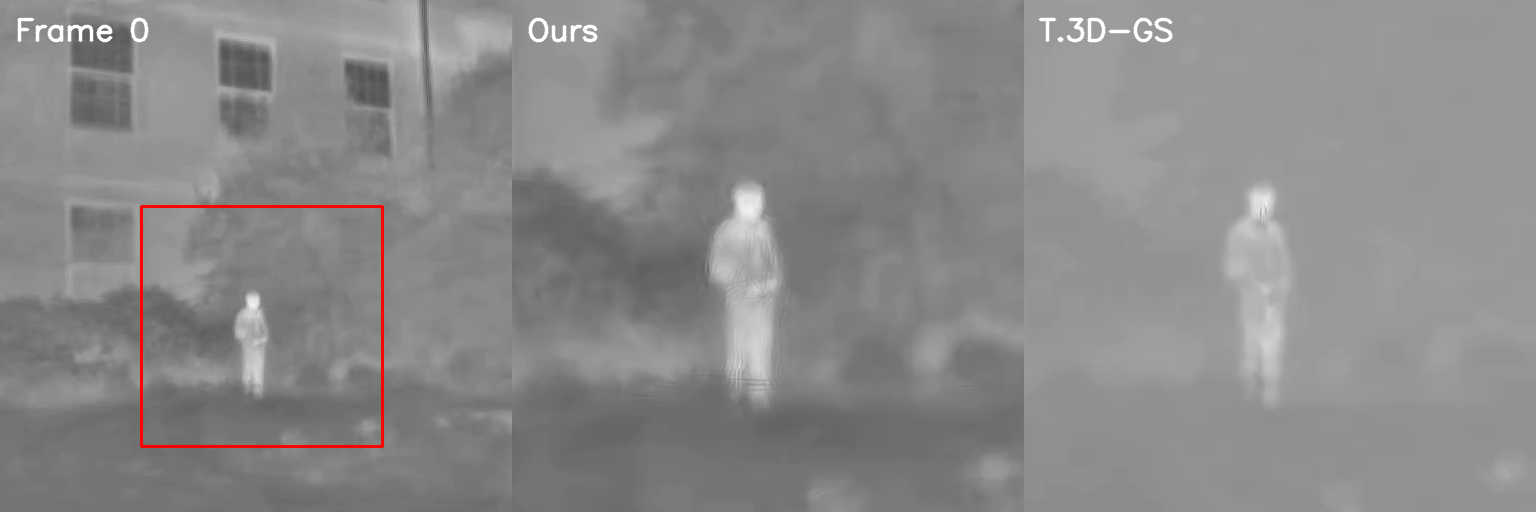}
   \caption{}
   \stepcounter{fi}
\end{figure}
\begin{figure}[ht!]
  \centering
  \includegraphics[width=\linewidth]{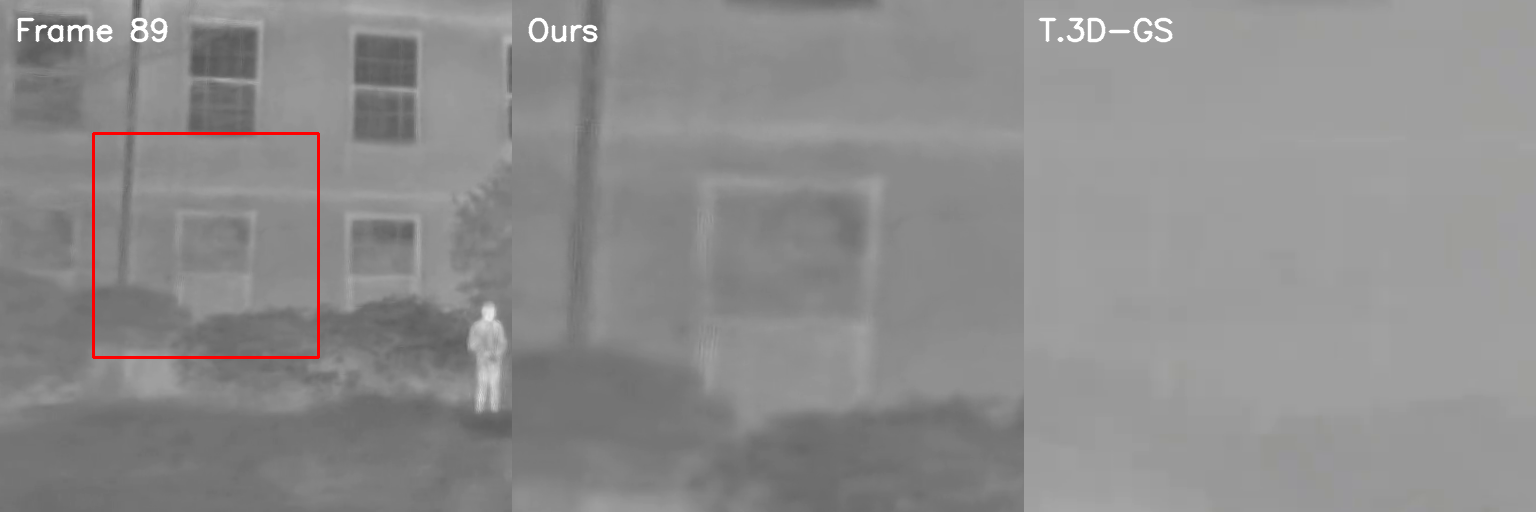}
   \caption{}
   \stepcounter{fi}
\end{figure}
\FloatBarrier

\pagebreak
\subsection{Lin et al. - Sink}
\begin{figure}[ht!]
  \centering
  \includegraphics[width=\linewidth]{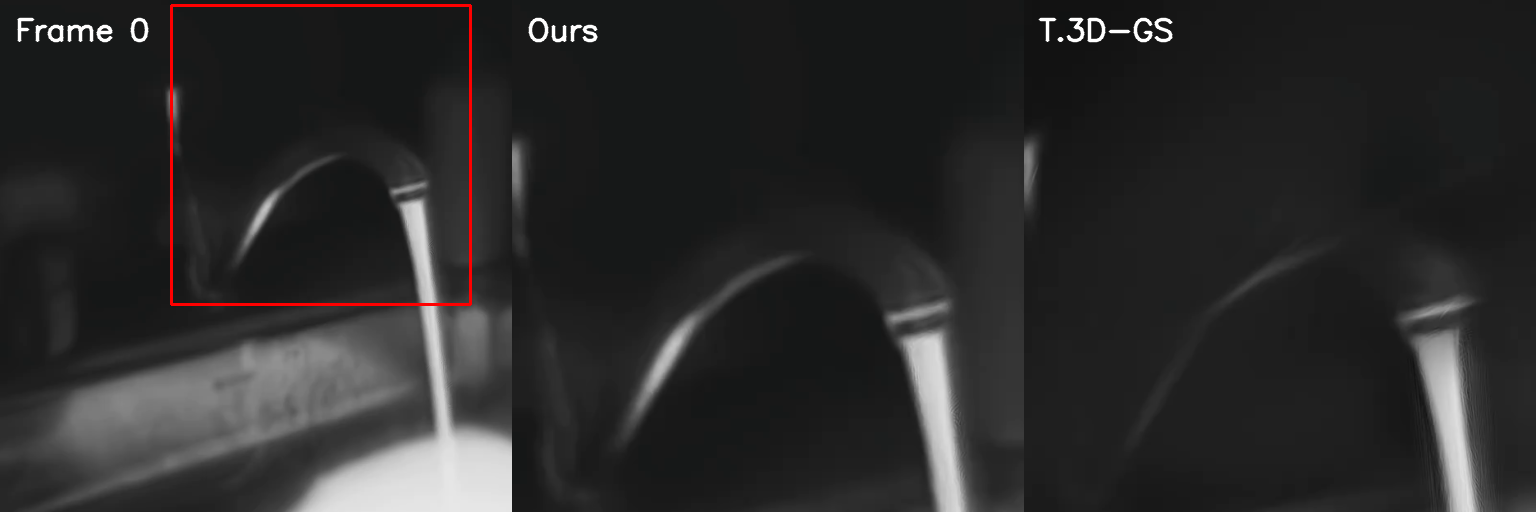}
   \caption{}
   \stepcounter{fi}
\end{figure}
\begin{figure}[ht!]
  \centering
  \includegraphics[width=\linewidth]{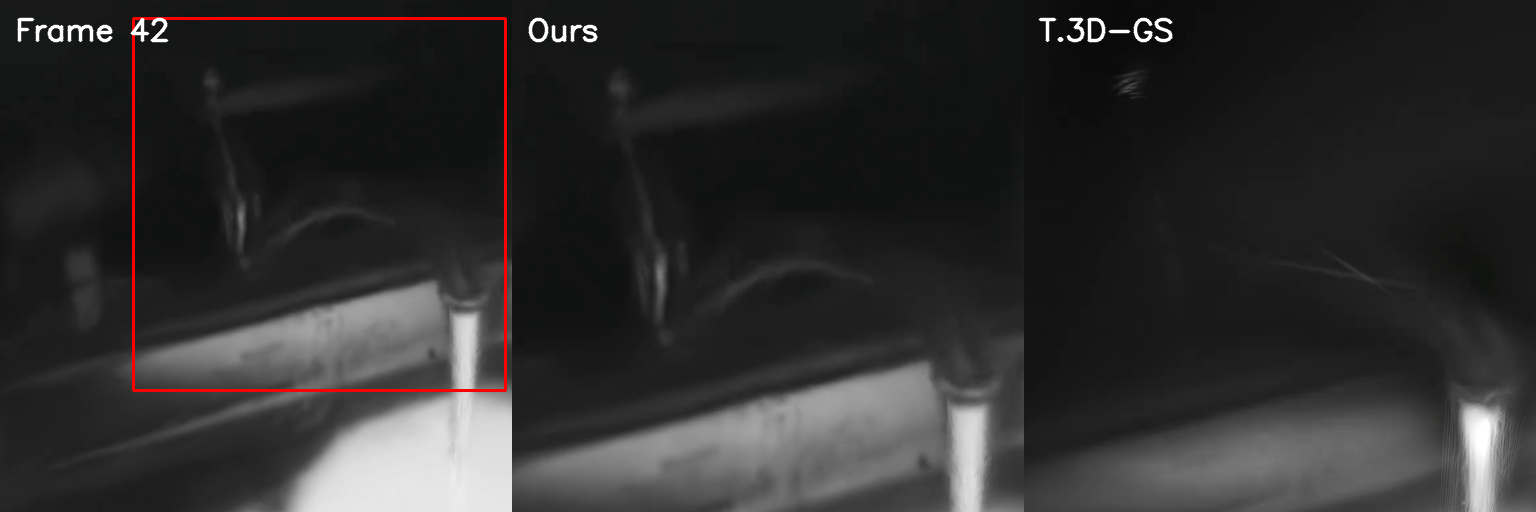}
   \caption{}
   \stepcounter{fi}
\end{figure}
\FloatBarrier

\subsection{MVTV - Mason}
\begin{figure}[ht!]
  \centering
  \includegraphics[width=\linewidth]{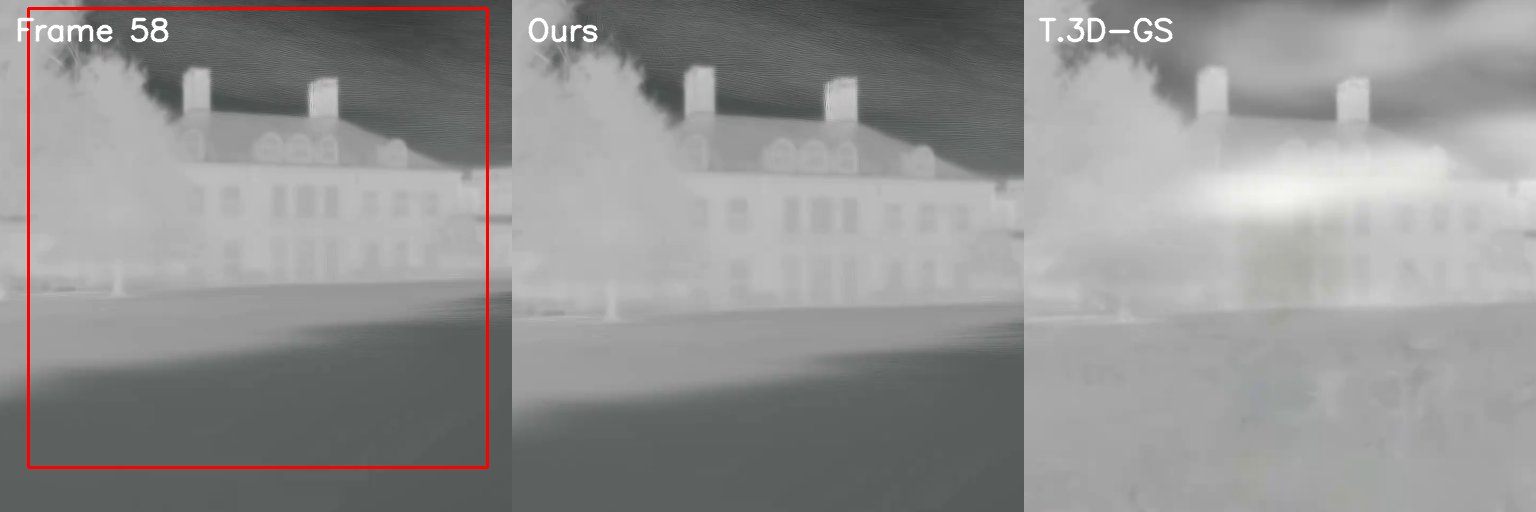}
   \caption{Floaters}
   \stepcounter{fi}
\end{figure}
\FloatBarrier

\pagebreak
\subsection{MSX - Ebike}
\begin{figure}[ht!]
  \centering
  \includegraphics[width=\linewidth]{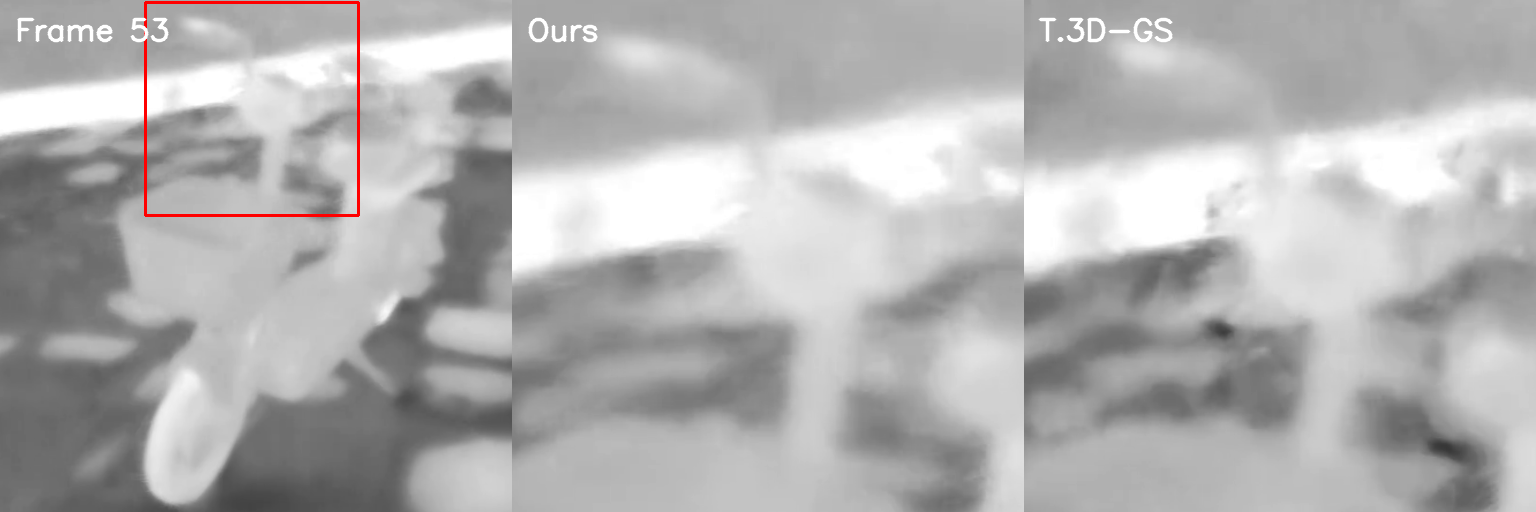}
   \caption{}
   \stepcounter{fi}
\end{figure}
\begin{figure}[ht!]
  \centering
  \includegraphics[width=\linewidth]{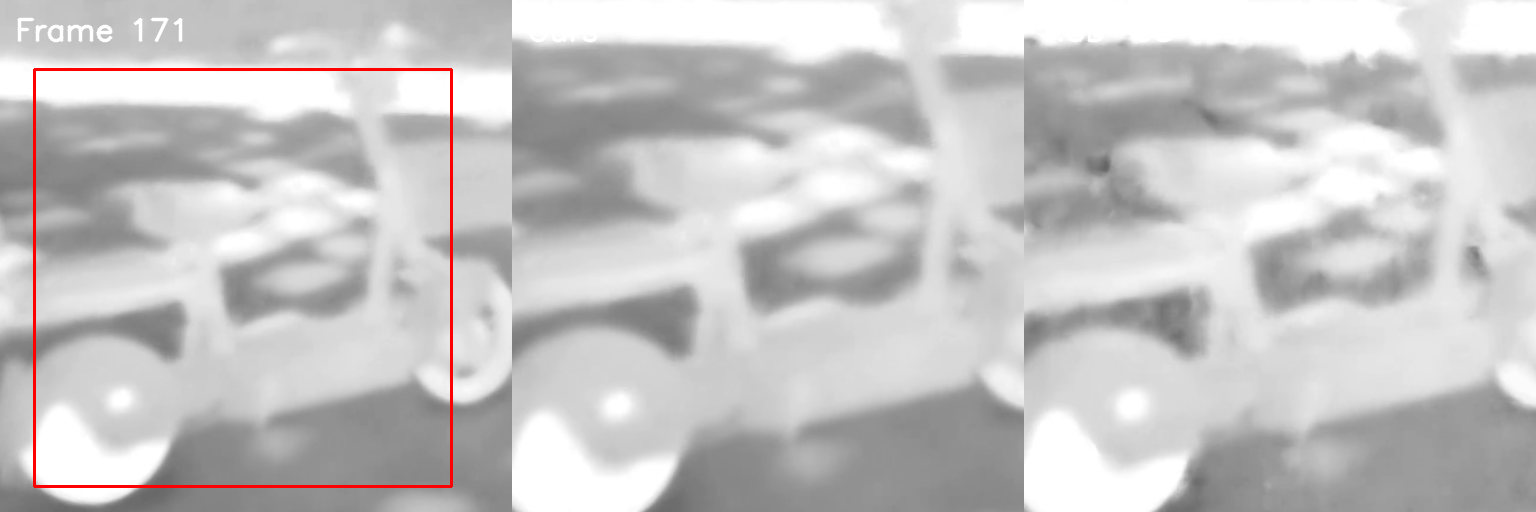}
   \caption{}
   \stepcounter{fi}
\end{figure}
\FloatBarrier

\subsection{ThermalMix - Face}
\begin{figure}[ht!]
  \centering
  \includegraphics[width=\linewidth]{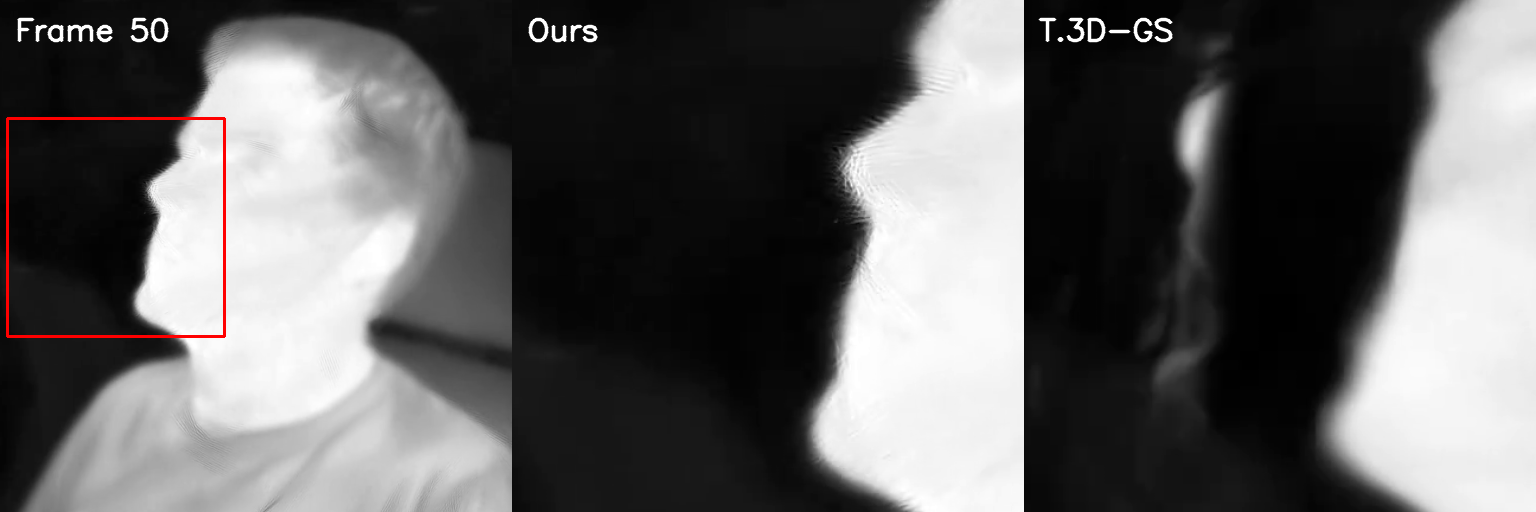}
   \caption{}
   \stepcounter{fi}
\end{figure}
\FloatBarrier

\subsection{MVTV - Chair}
\begin{figure}[ht!]
  \centering
  \includegraphics[width=\linewidth]{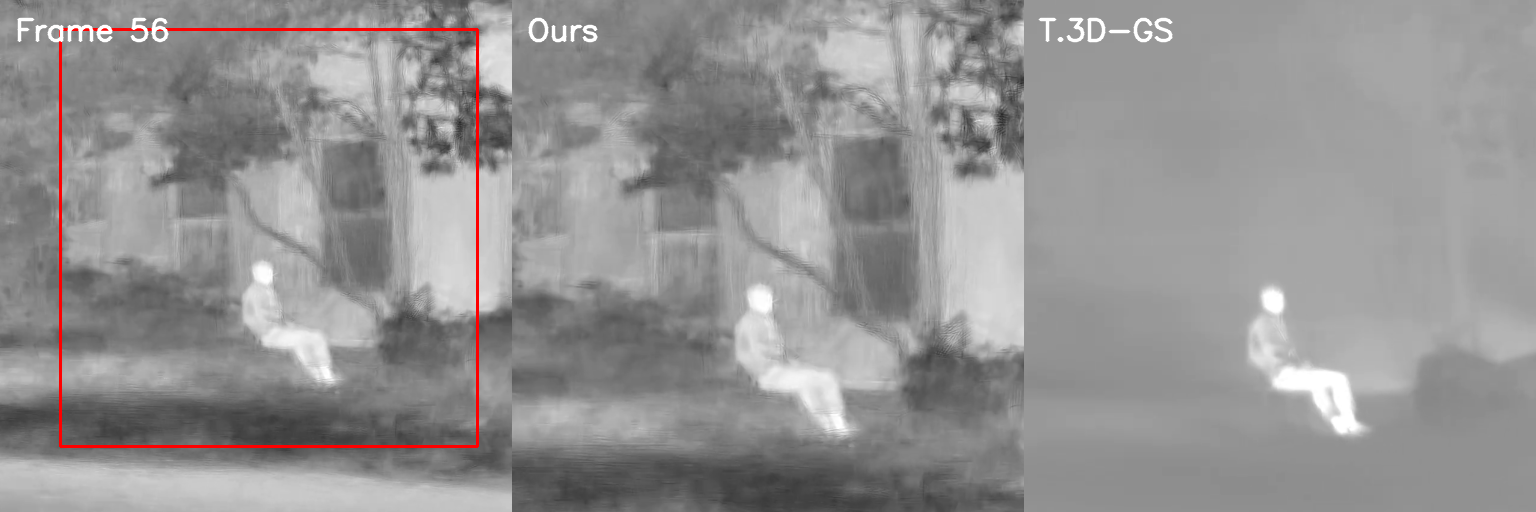}
   \caption{}
   \stepcounter{fi}
\end{figure}
\FloatBarrier

\pagebreak
\subsection{MSX - Building}
\begin{figure}[ht!]
  \centering
  \includegraphics[width=\linewidth]{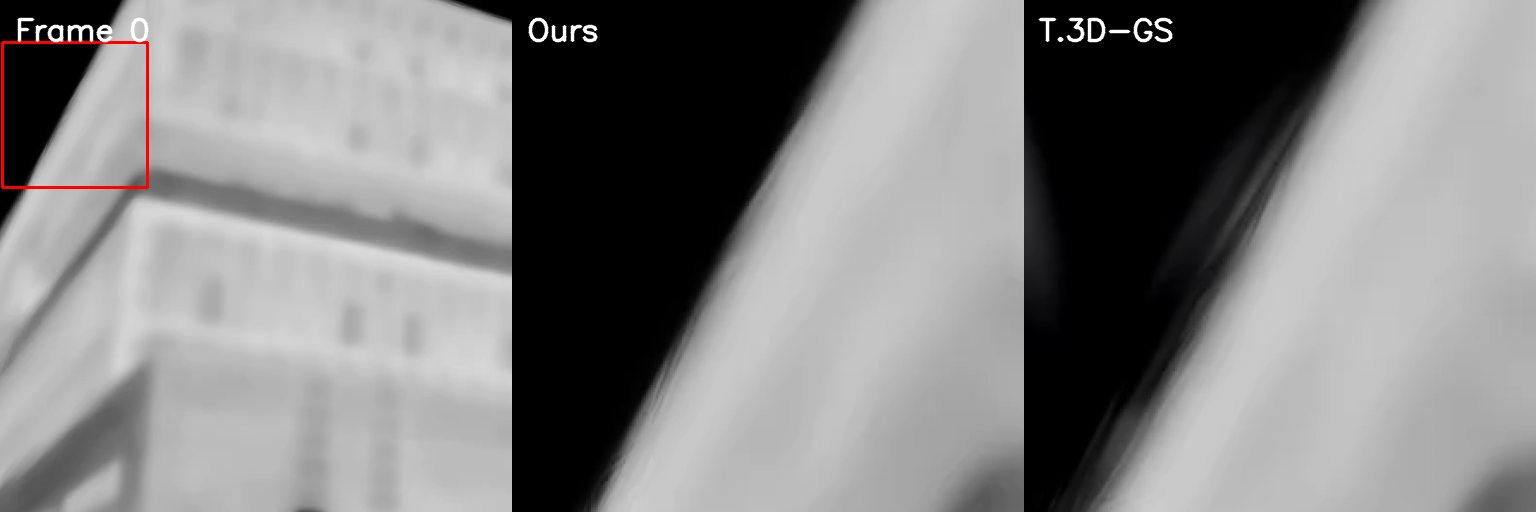}
   \caption{}
   \stepcounter{fi}
\end{figure}
\begin{figure}[ht!]
  \centering
  \includegraphics[width=\linewidth]{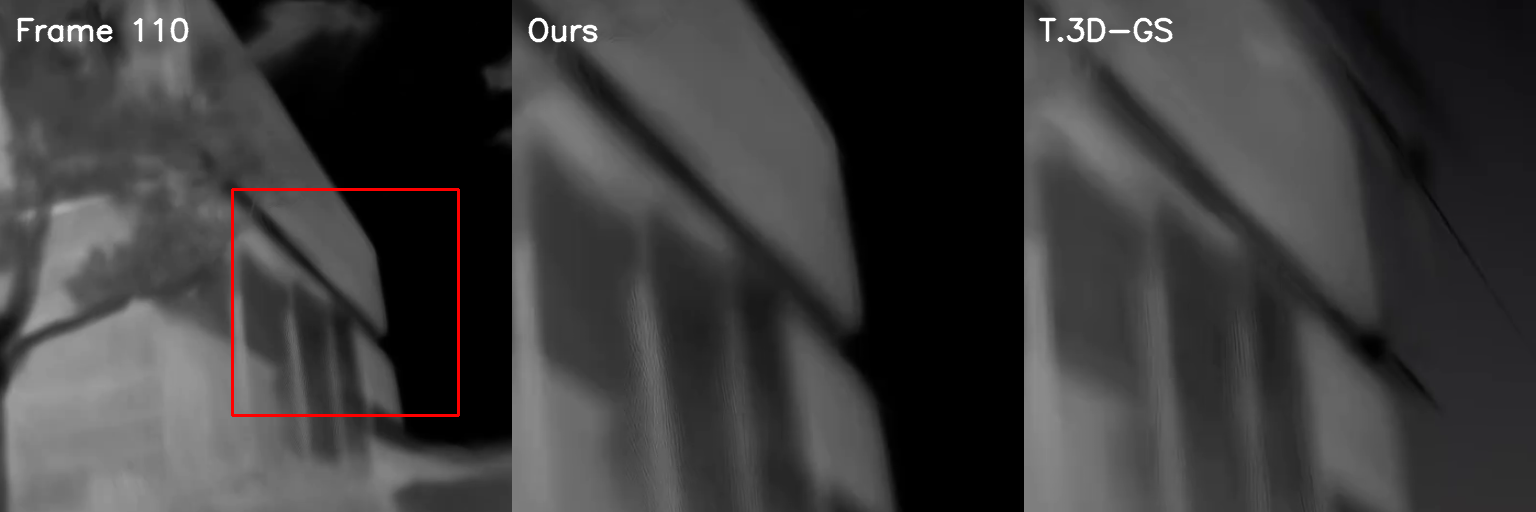}
   \caption{}
   \stepcounter{fi}
\end{figure}
\begin{figure}[ht!]
  \centering
  \includegraphics[width=\linewidth]{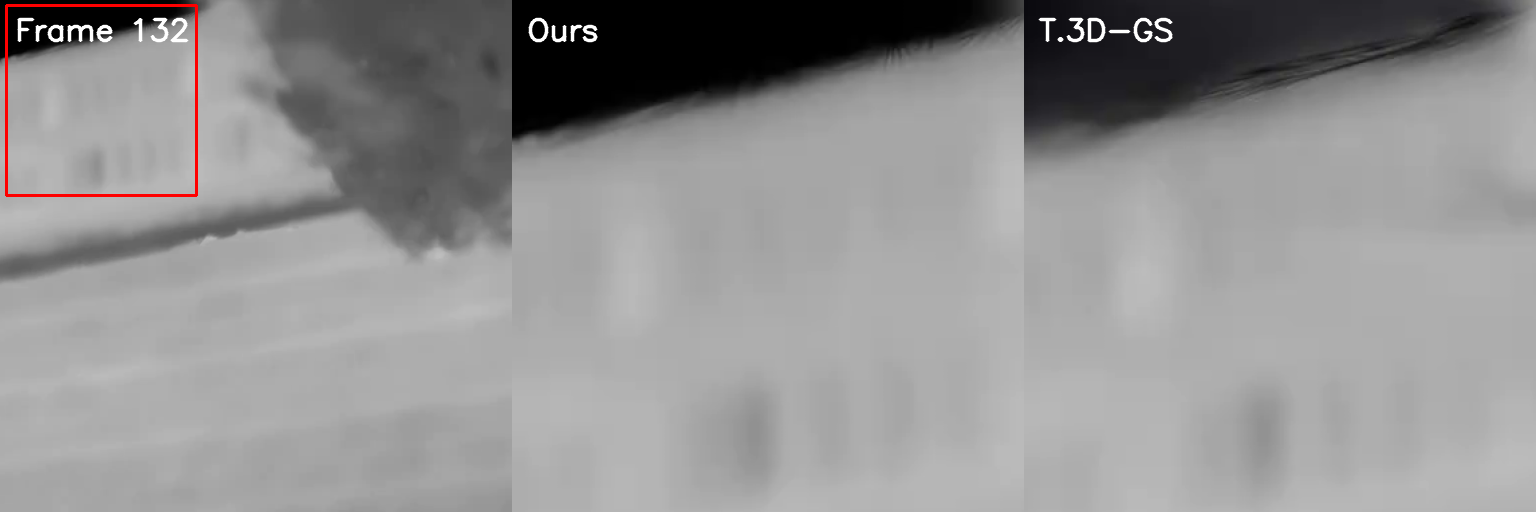}
   \caption{}
   \stepcounter{fi}
\end{figure}
\FloatBarrier

\subsection{MSX - Truck}
\begin{figure}[ht!]
  \centering
  \includegraphics[width=\linewidth]{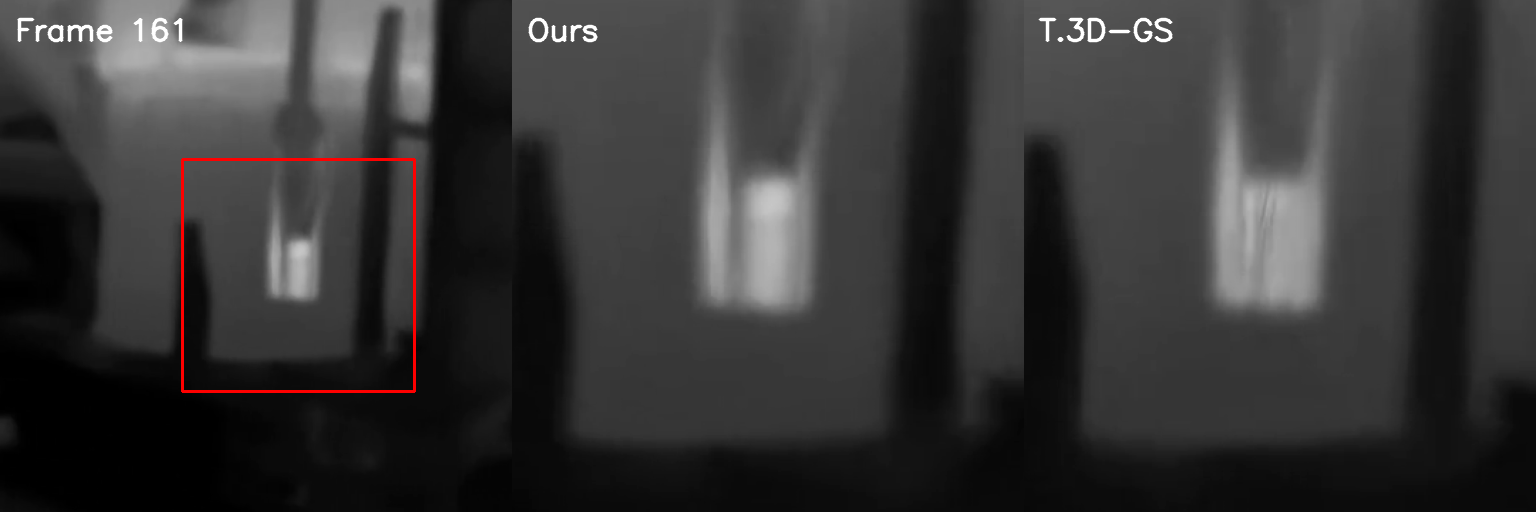}
   \caption{}
   \stepcounter{fi}
\end{figure}
\FloatBarrier

\pagebreak
\subsection{Lin et al. - Engine}
\begin{figure}[ht!]
  \centering
  \includegraphics[width=\linewidth]{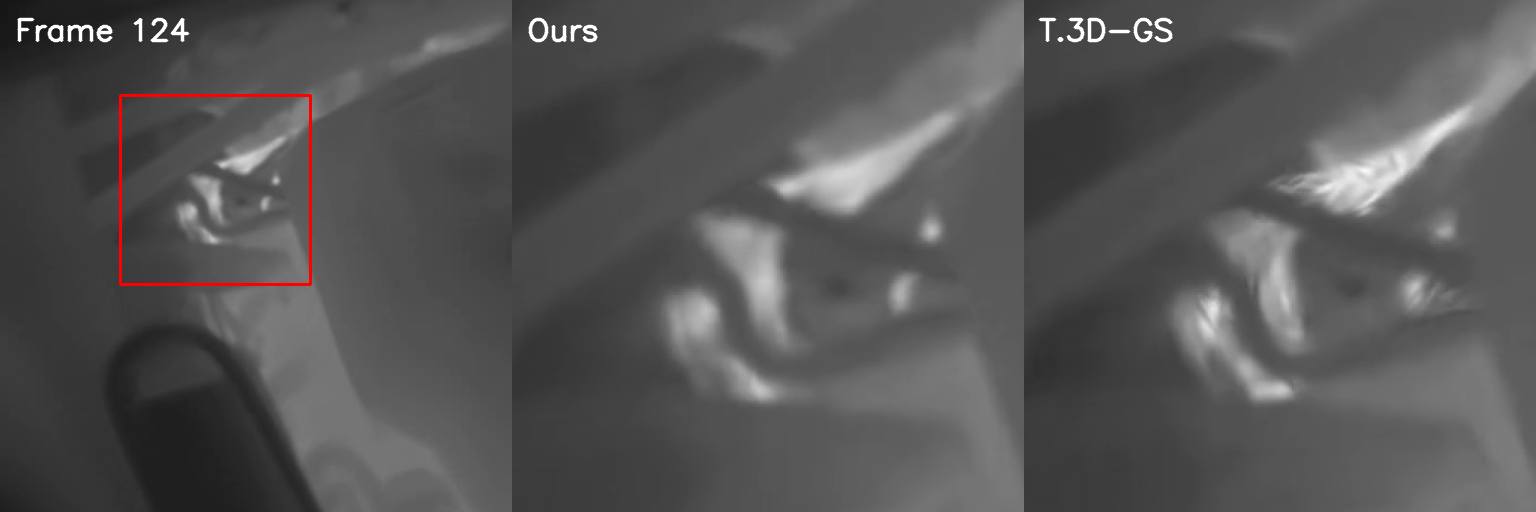}
   \caption{}
   \stepcounter{fi}
\end{figure}
\begin{figure}[ht!]
  \centering
  \includegraphics[width=\linewidth]{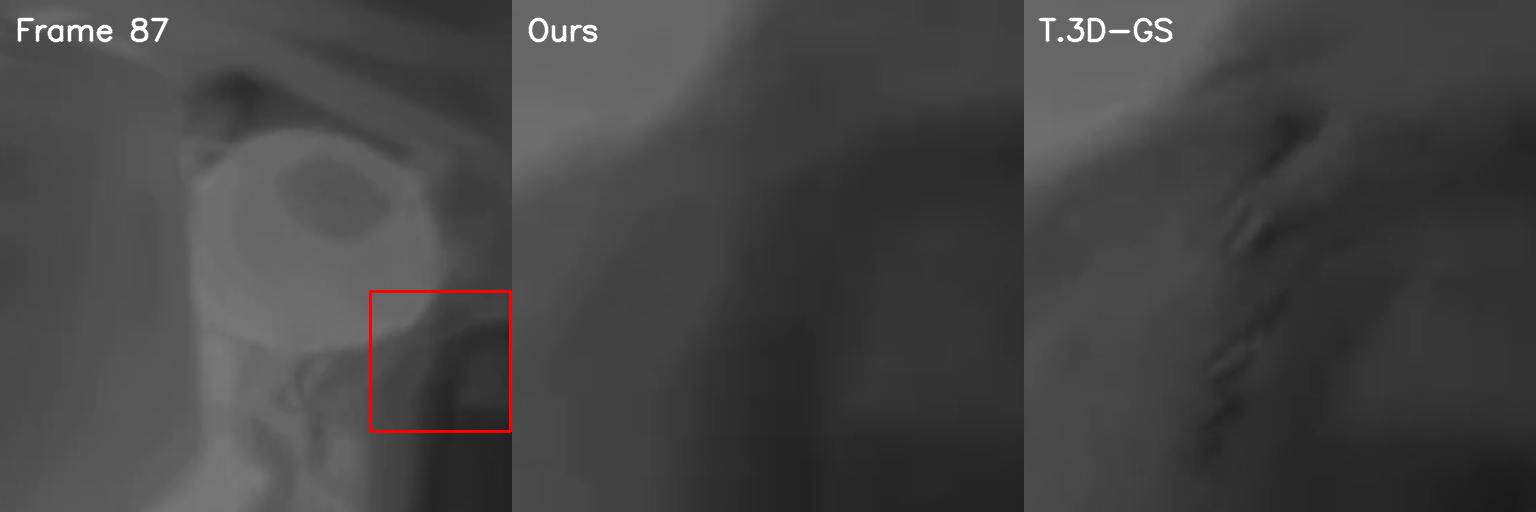}
   \caption{}
   \stepcounter{fi}
\end{figure}
\FloatBarrier

\subsection{MVTV - Parking}
\begin{figure}[ht!]
  \centering
  \includegraphics[width=\linewidth]{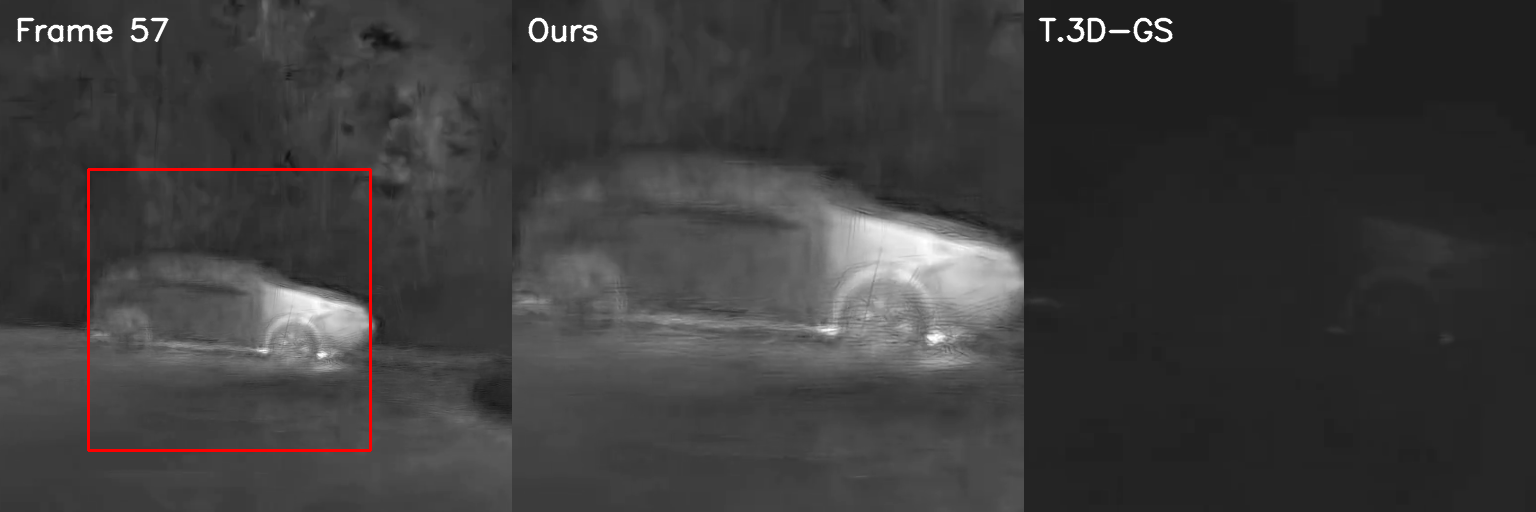}
   \caption{}
   \stepcounter{fi}
\end{figure}
\FloatBarrier

\newpage
\subsection{ThermalMix - Lion}
In this video, we crop the reconstructed volume around the lion using a small cubic region to better examine its structure. Without restricting the viewport, the NeRF results contain a large number of floaters. After cropping the volume, it becomes clear that there is no actual density corresponding to the lion’s geometry—only view-dependent floaters that explain the training images. This highlights the tendency of NeRF-based methods to overfit thermal inputs and fail to recover the actual geometry.

\begin{figure}[ht!]
  \centering
  \includegraphics[width=\linewidth]{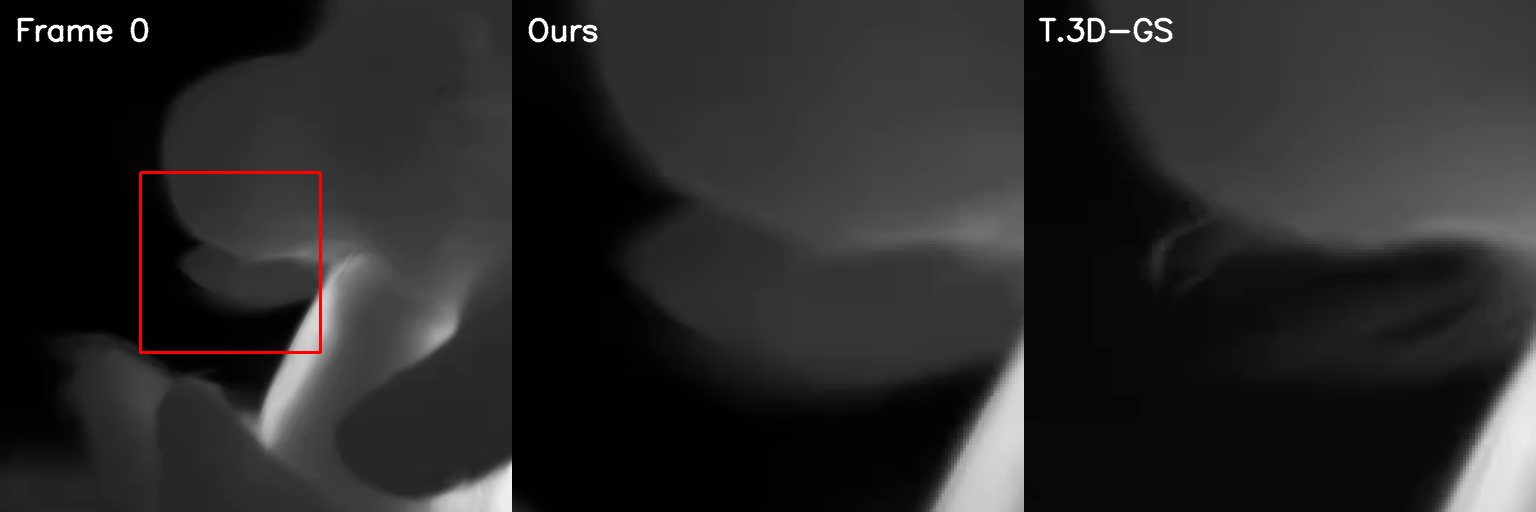}
   \caption{}
   \stepcounter{fi}
\end{figure}
\begin{figure}[ht!]
  \centering
  \includegraphics[width=\linewidth]{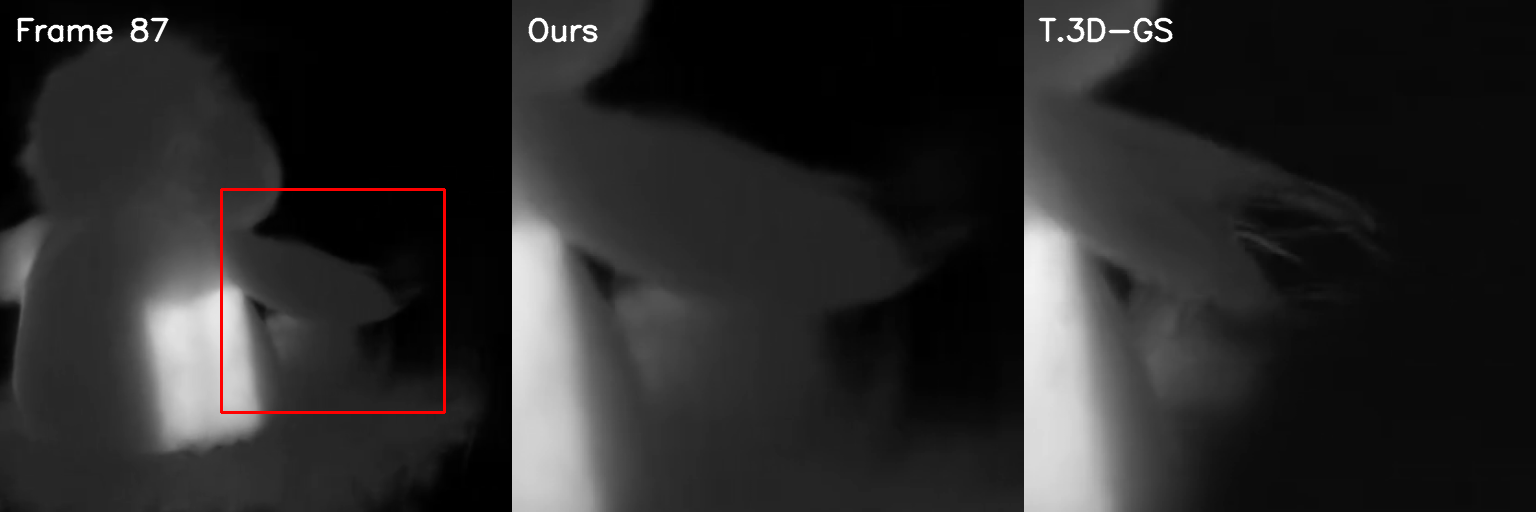}
   \caption{}
   \stepcounter{fi}
\end{figure}
\begin{figure}[ht!]
  \centering
  \includegraphics[width=\linewidth]{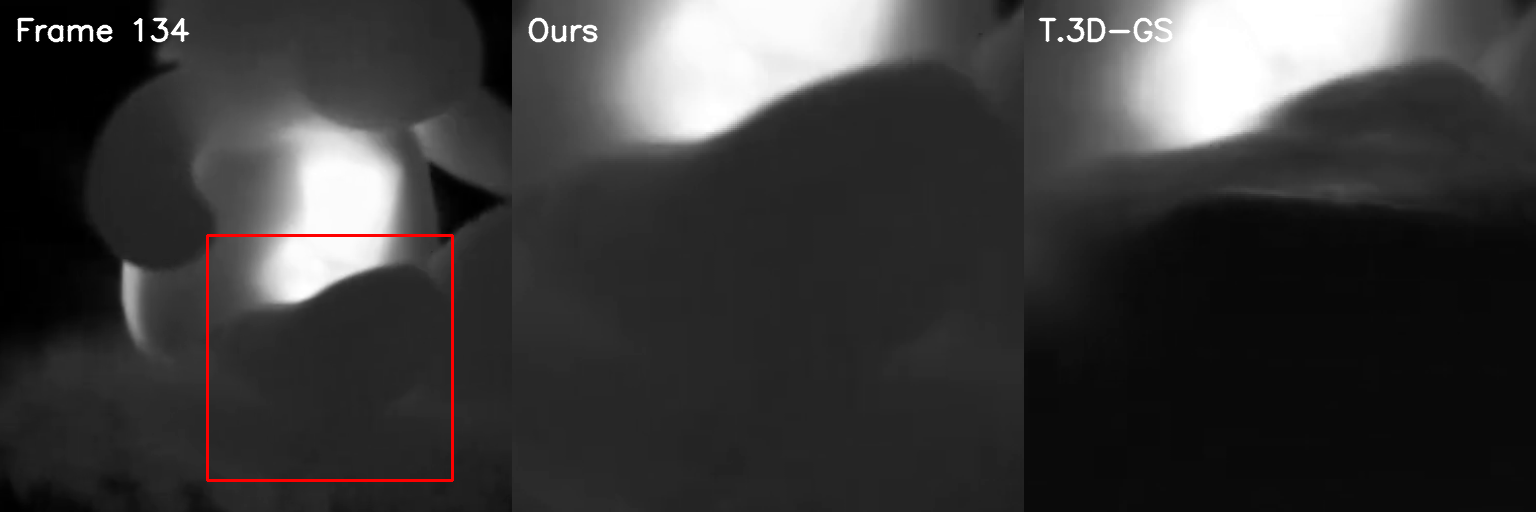}
   \caption{}
   \stepcounter{fi}
\end{figure}
\FloatBarrier
\stepcounter{si}

\end{document}